\documentclass[sn-mathphys]{sn-jnl}

\jyear{2022}%

\usepackage{overpic}
\theoremstyle{thmstyleone}%
\theoremstyle{thmstyletwo}%

\theoremstyle{thmstylethree}%

\newcommand{\figref}[1]{Fig.~\ref{#1}}%
\newcommand{\tabref}[1]{Table~\ref{#1}}%

\renewcommand{\eqref}[1]{Eq.~(\ref{#1})}
\begin{document}

\title[CASIA-Iris-Africa: A Large-scale African Iris Image Database]{ CASIA-Iris-Africa: A Large-scale African Iris Image Database }


\author[1,2]{\fnm{Jawad} \sur{Muhammad}}

\author[1,2]{\fnm{Yunlong} \sur{Wang}}

\author[1,2]{\fnm{Junxing} \sur{Hu}}

\author[1,2]{\fnm{Kunbo} \sur{Zhang}}

\author[1,2]{\fnm{Zhenan} \sur{Sun}}


\affil[1]{\orgdiv{School of Artificial Intelligence}, \orgname{University of Chinese Academy of Sciences}, \orgaddress{\city{Beijing} \postcode{100049},  \country{China}}}

\affil[2]{\orgdiv{CRIPAC, NLPR}, \orgname{Institute of Automation, Chinese Academy of Sciences}, \orgaddress{\city{Beijing} \postcode{100190},  \country{China}}}


\abstract{Iris biometrics is a phenotypic biometric trait that has proven to be agnostic to human natural physiological changes. Research on iris biometrics has progressed tremendously, partly due to publicly available iris databases. Various databases have been available to researchers that address pressing iris biometric challenges such as constraint, mobile, multispectral, synthetics, long-distance, contact lenses, liveness detection, etc. However,  these databases mostly contain subjects of Caucasian and Asian docents with very few Africans. Despite many investigative studies on racial bias in face biometrics, very few studies on iris biometrics have been published, mainly due to the lack of racially diverse large-scale databases containing sufficient iris samples of Africans in the public domain. Furthermore, most of these databases contain a relatively small number of subjects and labelled images. This paper proposes a large-scale African database named CASIA-Iris-Africa that can be used as a complementary database for the iris recognition community to mediate the effect of racial biases on Africans. The database contains 28,717 images of 1023 African subjects (2046 iris classes) with age, gender, and ethnicity attributes that can be useful in demographically sensitive studies of Africans. Sets of specific application protocols are incorporated with the database to ensure the database's variability and scalability. Performance results of some open-source SOTA algorithms on the database are presented, which will serve as baseline performances. The relatively poor performances of the baseline algorithms on the proposed database despite better performance on other databases prove that racial biases exist in these iris recognition algorithms. The database will be made available on our website: http://www.idealtest.org.}




\maketitle

\section{Introduction}\label{sec1}

Iris is a biometric trait with phenotypic characteristics that has proven to be agnostic to human natural physiological changes such as aging \cite{2008Image}. Recently, there have been tremendous achievements in the research, development, and deployment of iris biometric systems over a wide range of applications such as in security, access control, surveillance, law enforcement, banking, education, etc. These systems have critically evolved in the past two decades partly due to the availability of public iris databases. \cite{omelina2021survey}.

Various forms of iris databases had been published that address some specific research problems and are made available to the biometrics research community such as constraint databases \cite{casia_v2}, that provide use cases for performing fundamental research on iris variability and repeatability of results performance; mobile iris \cite{de2015mobile}, \cite{VISOB_Dataset}, that made possible to simulate iris biometric performance in the fast growing mobile environment; multispectral \cite{crossEyed7736915}, \cite{Toward2616281}, which provide benchmarks for robust evaluation of iris recognition in a dual spectrum paradigm that can lead to enhanced performance; synthetics \cite{WVU4100635}, which provides a large-scale iris database with various degrees of controlled imaging conditions that can be used to robustly evaluate iris recognition algorithms; iris at a distance \cite{casia_v4}, \cite{UBIRISv2}, which present real-time iris images captured by less user cooperation with challenging conditions like the varying iris size, image reflections, occlusion, blur, low resolution, off-axis, motion, specular reflections, etc that can be used to simulate real-time iris recognition performance; contact lenses \cite{contact6776569}, special case iris databases captured with contact lenses that provide the unique opportunity for algorithms to evaluate performances on irises covered by different types of contact lenses; liveness detection \cite{liveness8272763}, set of databases that enable the critical examinations of algorithms security against spoofing attacks; Aging \cite{aging6239214}, proposed to be specially captured over a long period of a time interval of multiple years which can be used for studying the known iris biometric feature of aging.

\begin{figure}
  \centering 
  \begin{overpic}[ height=7cm]{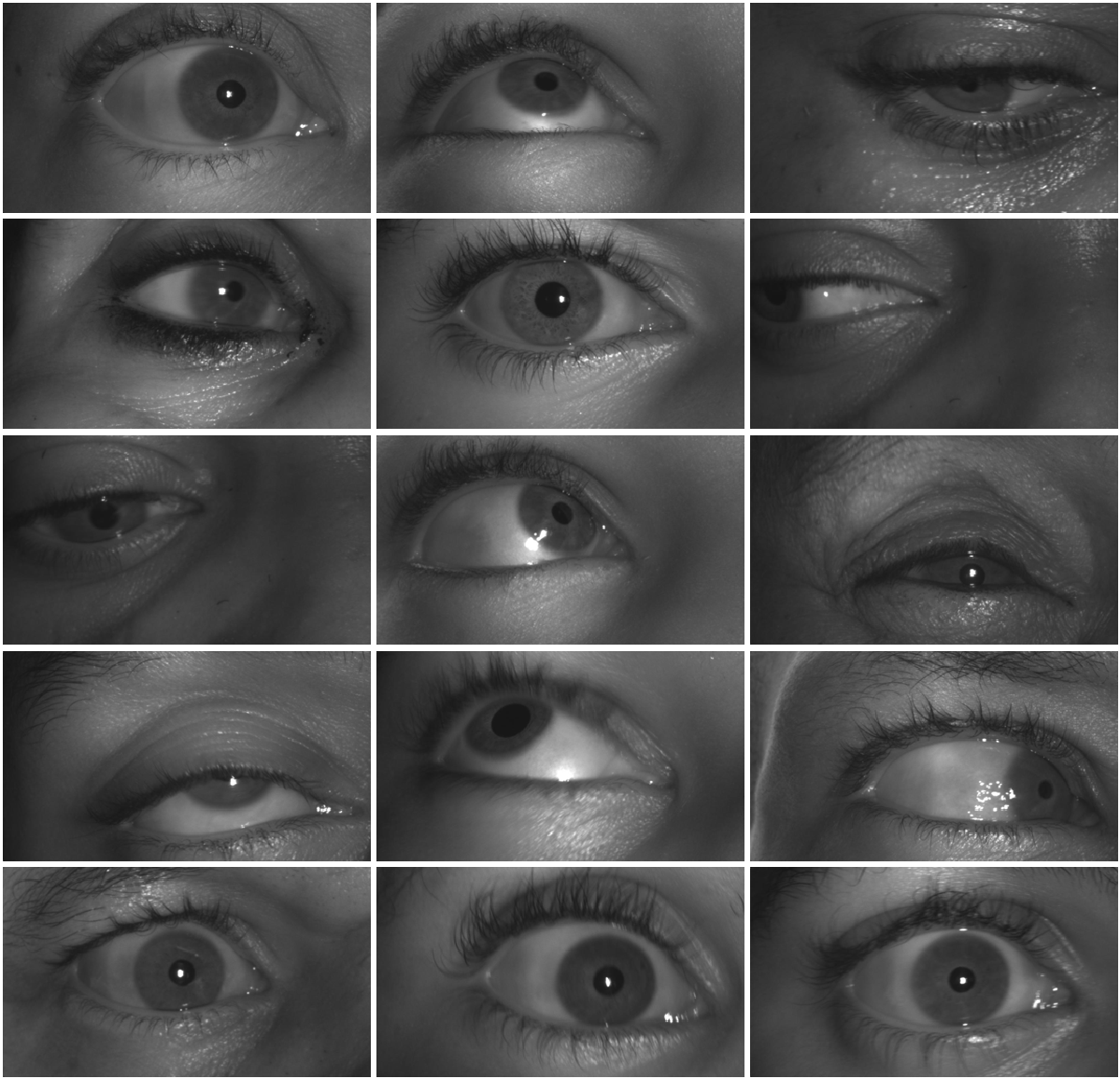}
  \end{overpic}
  \caption{ CASIA-Iris-Africa sample images  }
 \label{fig:highlight_image}\vspace{-3pt}
\end{figure}

However,  even though these databases have addressed many of the problems mentioned above, most of them contain subjects of primarily Caucasian and Asian docents with very few Africans and many times zero Africans as outlined in the iris databases summary of \tabref{tab:iris_databases}. It is particularly challenging as this has created a research blind spot for African cohorts in these databases. Despite many of the reported investigative studies on racial bias in face biometrics \cite{Patrick2019Face}, \cite{Buolamwini1}, \cite{cavazos2020accuracy}, very few iris racial bias-related studies have been published, which can be due to the insufficient number of other races databases such as the Africans. Recently, face recognition algorithms have been reported to be biased toward specific demographics. Most prominently, many investigative studies have reported higher false-positive rates in African subject cohorts than in other cohorts. Due to the unavailability of the required quantity of African cohorts in the publicly available iris databases, similar studies would be difficult to replicate on iris biometrics. This paper presents a mainly African dataset (samples shown in \figref{fig:highlight_image}) that can be useful in addressing some of these challenges.
Specifically:

\begin{itemize}
  \item A large-scale African iris image database named \emph{CASIA-Iris-Africa} is presented and will be made available to the iris biometric research community. The database is captured from 1,023 subjects and has 2046 unique irises with 28,717 images. It can be used as a complementary dataset to the existing databases and helps spur studies on iris biometrics of African-related racial bias. To our knowledge, this database is the first of its publicly available kind.
  \item Sets of application protocols have been developed that will objectively ensure performance variability and repeatability of iris recognition algorithms on the database. These protocols cover potential database application scenarios in iris identification, verification, and classification.
  \item The performance results of various SOTA algorithms on the proposed database protocols are presented, which can be used as the database baseline for further research and provide preliminary insights into racial bias against Africans in iris biometrics.
\end{itemize}

The remainder of this paper is organized as follows: In section II, related work in the literature will be discussed. Section III describes the proposed database in detail, and the baseline evaluation performance of the database is provided in section IV. Section V concludes the paper and suggests future work.

\section{Related Work}
\label{sec::related_Work}

\subsection{Iris image databases}
In \tabref{tab:iris_databases}, a summary of some of the most popular iris databases with their race distribution is presented. It can be observed that there are very few subjects in the African cohort. Although many of these databases did not report their race distribution, it is safe to assume that the unreported subjects are not of African descent, considering the origin of these databases, which are mostly acquired from their local communities. Hence, a large-scale African database such as the proposed CASIA-Iris-Africa is necessary to complement these existing databases.

\begin{table}
  \begin{center}
  \begin{minipage}{\textwidth}
  \renewcommand{\arraystretch}{1.1}
  \setlength\tabcolsep{3pt}
  \caption{Some popular Iris biometrics databases and their racial distribution.} \label{tab:iris_databases}%
  
  \footnotesize
  \begin{tabular*}{\textwidth}{ p{0.8\linewidth}p{0.15\linewidth}}
    \toprule
Database  Description &   Race
   \\ \hline

\textbf{ND-IRIS-0405  \cite{bowyer2016nd}}:  Captured using the LG2200 sensor. Some of the images in the database correspond to subjects wearing contact lenses, leading to visible image artifacts. Furthermore, the database has a roughly balanced female-male ratio.   \emph{Subjects:} 712; \emph{Classes:} 56; \emph{Images: }  64,980 (NIR) &   AS (23\%); C (70.3\%); O (6.7\%)
\\ \hline

\textbf{UBIRIS-v1  \cite{proencca2005ubiris} }: Captured using Nikon E5700 comprises images with various degrees of reflections, contrast, luminosity, and focus problems. \emph{Subjects:} 482; \emph{Classes:} 241; \emph{Images: } 1877( VIS) & -
\\ \hline

 \textbf{UBIRIS-v2  \cite{UBIRISv2} }: Captured using Canon EOS 5D DSLR sensor and contains multisession iris images that singularly contain data captured in the visible wavelength, at-a-distance (between four and eight meters), and on the move. \emph{Subjects:} 522; \emph{Classes:}  261; \emph{Images: } 11,102 (VIS) & AS (2\%); C (90\%); A (8\%)
\\ \hline

\textbf{IIT Delhi (IITD-V1) \cite{KUMAR20101016}\cite{IITDWebsite} }:  Captured using IRIS, JPC1000, and digital CMOS camera from the students and staff at IIT Delhi, New Delhi, India. \emph{Subjects:} 224; \emph{Images: } 1120 (NIR) & AS-S (100\%)
    \\ \hline

 \textbf{ICE 2005 \cite{phillips2008iris} }:  Captured using LG EOU 2200 and comprises images with a broader range of quality than the sensor would normally acquire, including low-quality images that the sensor will otherwise reject.  \emph{Subjects:} 244; \emph{Classes:}  112; \emph{Images: } 2953 (NIR) & -
    \\ \hline

\textbf{MMU-v1 \cite{mmu_db} }:  From the Multimedia University Iris Database, captured by the LG IrisAccess sensor. The Iris images introduce eyelash obstruction and eye rotation. \emph{Subjects:} 90; \emph{Classes:}  45; \emph{Images: } 450(NIR) & -
    \\ \hline

 \textbf{MMU-v2 \cite{mmu_db} }:  Captured by Panasonic BM-ET100US Authenticam sensor. This is another version of MMU-v1 introduced after one year.  \emph{Subjects:} 200; \emph{Classes:}  100; \emph{Images: } 995(NIR) & -
    \\ \hline

 \textbf{MICHE-I \cite{de2015mobile} }: Captured by the volunteers using their mobile devices: iPhone5, Galaxy Samsung IV (GS4), and Galaxy Tablet II (GT2) on their hands. No restriction was imposed on subjects as they chose any face accessory as they would in real practical application.  \emph{Subjects:} 184; \emph{Classes:}  92; \emph{Images: } 3732(VIS) & C (100\%)
    \\ \hline


 \textbf{CROSS-EYED -\cite{crossEyed7736915} }:  This Dual spectrum database contains iris and periocular images captured at a distance of 1.5 meters using a custom multispectral iris sensor under an uncontrolled realistic indoor environment and challenging illumination reflection.  \emph{Subjects:} 240; \emph{Classes:}  120; \emph{Images: } 11,520(NIR/VIS) & C (75\%); AS (17\%); AS-S (4\%); A (3\%)
    \\ \hline

 \textbf{Aging Analysis \cite{aging6239214} }:  Private dataset that was captured by LG 400 from 2008 through 2011  and used for a comprehensive aging analysis in iris biometrics. \emph{Subjects:} 644; \emph{Classes:}  322; \emph{Images: } 22,156(NIR) & C (76\%); AS (12\%); O (12\%);
    \\ \hline

 \textbf{CASIA IrisV1 \cite{Ma02irisrecognition} }:   Captured with a custom digital optical sensor over two sessions one month apart.  \emph{Subjects:} 50; \emph{Classes:}  25; \emph{Images: } 500 (NIR) & AS (100\%)
    \\ \hline

 \textbf{CASIA IrisV2 \cite{casia_v2} }:   Iris images were captured using two hand-held iris devices (OKI IRISPASS-h and CASIA-IrisCamV2) in an indoor setup over a single session.  \emph{Subjects:} 60; \emph{Images: } 2400 (NIR) & AS (100\%)
   \\ \hline


 \textbf{CASIA IrisV3-Lamp \cite{casia_v3} }: Captured using the handheld OKI IRISPASS-h iris device and a Lamp light source turned on/off close to the subject to introduce more intraclass variations. As such, the captured irises have nonlinear deformation due to variations in visible illumination. \emph{Subjects:} 819; \emph{Classes:}  411; \emph{Images: } 16,212 (NIR) & A (100\%)
    \\ \hline


 \textbf{CASIA IrisV4-Thousand \cite{casia_v4} }: Captured using IKEMB-100 camera from IrisKing. The database contains iris images with eyeglasses and specular reflections and provides a wide range of intraclass variations.  \emph{Subjects:} 2000; \emph{Classes:}  1000; \emph{Images: } 20, 000 (NIR) & AS (100\%)
    \\ \hline
\textbf{CASIA-Iris-Africa }: Our proposed database is captured using an IKUSBE30 iris sensor with varying degrees of iris sizes across inter and intraclasses.  \emph{Subjects:} 2046; \emph{Classes:}  1023; \emph{Images: } 28,717 (NIR) & A (100\%)
    \\ \hline

 \bottomrule
 
 \end{tabular*}
 
 \footnotetext{A: Black-or-African-American; 	C: White/Caucasian; 	AS: Asian; 	AS-ME: Asian-Middle-Eastern; 	AS-S: Asian-Southern; 	O: Other. These race labels are loosely adopted from \cite{Phillips26} }

\end{minipage}
\end{center}
\end{table}

\subsection{Iris segmentation and localization}
Usually, iris biometrics is based on a constrained setup, whereby the iris images are captured under stringent configurations, producing qualitative iris images that simplify the iris preprocessing task of localization and segmentation. However, more recently, unconstrained iris biometrics, which require less user cooperation and can operate at larger distances, have become more prevalent. This attracts more user applications due to its relaxed invasiveness but also introduces numerous challenges such as occlusion, blur, low resolution, off-axis, motion, and specular reflections\cite{li2021robust}. Various methods have been proposed to address these challenges, which can be broadly categorized \cite{ahmad2018unconstrained} into specialized, hybrid, and deep learning methods. Specialized approaches propos the use of prior information (such as iris circular/elliptical shape and dark pupil region) that is specific to iris trait features and can be unambiguously identified in an iris image \cite{Daugman2009How},\cite{Tan2013Towards},\cite{Gangwar2016IrisSeg}. These methods are fast and do not require training images. They can be effective for loosely unconstrained iris images of high quality, as the handcrafted features would be easily identified. However, these methods would be seriously limited for completely unconstrained iris images as the handcrafted features would be less effective. The hybrid methods proposed an alternative approach that complements specialized methods with machine learning techniques to improve the effectiveness of the ILS \cite{Tan2012Unified},\cite{Radman2017Automated},\cite{Proena2010Iris}. Coarse segmentation is usually obtained with a general machine learning technique, and specialized method is applied to generate the final segmentation. The machine learning algorithms adopted by a hybrid method depend on the training process with iris images. Hence, hybrid methods employ iris priors and ground truth iris images. These methods can be more accurate but slower than specialized methods. However, since they still heavily rely on iris priors, these methods can also be ineffective in challenging unconstrained iris images. To mitigate the problem of unreliable iris priors, deep learning approaches \cite{ahmad2018unconstrained}, \cite{Sardar2020Iris},\cite{Wang2020Towards}, propose solutions based on general-purpose semantic segmentation methods. They are more accurate than the former methods and do not require crafting any prior iris features. However, they usually require large-scale training data and are generally slower.
In this work, some of the methods in these three categories have been adopted to generate benchmark results for the proposed database.

\subsection{Iris representation}

Gabor filter \cite{fogel1989gabor} based features are among the most popular classic iris representation methods \cite{Daugman2009How}, \cite{masek2003recognition}. In \cite{Daugman2009How}, Daugman proposed the application of a 2D Gabor filter on a normalized iris image, and the resulting responses were binarized to produce IrisCodes. Then the Hamming distance was used to measure the similarity between two IrisCodes. In \cite{masek2003recognition}, a 1D Gabor filter was used instead of a 2D filter to efficiently generate  IrisCodes. In \cite{sun2008ordinal}, rather than the precise measurement of the image structures of the Gabor filter-based features, qualitative relationships between the iris image regions were instead measured as ordinal features using multilobe differential filters. Another classic approach is the use of discrete cosine transforms(DCT) \cite{ahmed1974discrete} to generate binary features by analysing the frequency of image blocks in a normalized iris image \cite{monro2007dct}. In \cite{miyazawa2008effective}, discrete Fourier transforms(DFT) \cite{sundararajan2001discrete} were employed instead of the DCT to generate the binary features from the frequency of the image blocks.

Beyond the classical methods of iris representation, deep learning-based approaches for extracting the iris features in an image are becoming more prevalent \cite{gangwar2016deepirisnet}, \cite{zhao2017towards}, \cite{zhang2018deep}. In \cite{gangwar2016deepirisnet}, a convolutional neural network (CNN) trained with normalized iris images was employed with the output vector of the last fully connected layer considered as the iris features. In \cite{zhao2017towards}, instead of learning direct iris representation from the last layers, the CNN learned spatially corresponding features by stacking the output of multiple layers and convolving them to produce a single channel feature image which is then binarized to generate the IrisCodes. In \cite{zhang2018deep}, a method coined as MaxOut involves a weighted fusion of the CNN deep features of an iris and periocular biometric modality into a compact representation that can be used for iris recognition.

\section{The Database}
\label{sec::database}
\subsection{ Locations }
The iris images were captured in Nigeria, Africa. The capturing exercise was conducted across various locations in multiple sessions over three months. Due to logistical constraints, the capturing exercise was limited to the northern part of Nigeria. However, the locations were strategically selected within the region to improve the subjects' diversity. These locations are in urban cities, schools, and local neighborhoods. The locations are shown in \tabref{tab:capturing_locations}. Each subject volunteer was asked to sign a consent form authorizing the data to be used for only research purposes.
\begin{table*}[!htbp]
    \caption{Database Capturing Locations}
    \label{tab:capturing_locations}
    \centering
    \footnotesize
    \setlength{\tabcolsep}{4pt}
    \renewcommand{\arraystretch}{1.2}
    \begin{tabular}{p{0.3\linewidth}|p{0.6\linewidth}}
        \hline
        Location & Notes \\
        \hline
        Ahmadu Bello University Zaria, Kaduna state. &   A diverse population that comes from locations all over Nigeria across multiple ethnicities \\ \hline
        Kano State Polytechnic, Kano state. &   Diverse age demographics that could not be found elsewhere. \\ \hline
        Gandun Albasa street, Kano state. &   Availability of specific subjects volunteers that were very difficult to be convinced.  \\ \hline
        Dabai, Danja, Katsina state. &   Attract more diverse subjects of all demographics  \\ \hline
        Brigate street, Kano state. &   Specifically selected because of the availability of subjects of various ethicalities that could not be found elsewhere.  \\ \hline
        Sabon Gari, Kano state. &   Additionally, chosen because of population diversity  \\ \hline
        Hotoro, Kano state. &   Additionally, chosen because of population diversity  \\
        \hline
    \end{tabular}
\end{table*}

\subsection{ Image capture }
\begin{figure}
  \centering
  \begin{overpic}[height=4cm]{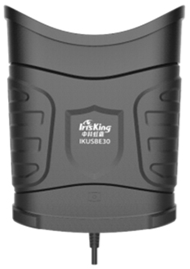}
  \end{overpic}
  \caption{ IKUSBE30 sensor used for the data capture  }
 \label{fig:iris_device}\vspace{-3pt}
\end{figure}

\begin{figure}
  \centering
  \begin{overpic}[height=5cm]{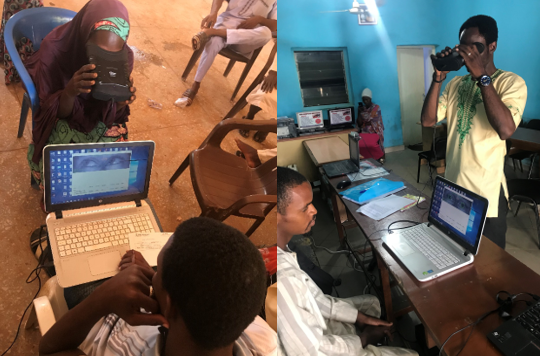}
  \end{overpic}
  \caption{ Iris Capturing set-ups (indoor and outdoor)  }
 \label{fig:capturing}\vspace{-3pt}
\end{figure}

The iris sensor device for capturing the images is the IKUSBE30 iris sensor from IrisKing shown in \figref{fig:iris_device}. The setup for the capturing exercise is shown in \figref{fig:capturing}. Each volunteer was asked to hold the device, standing or sitting, based on their preference. As such, the iris images were captured at various degrees of head postures.

\begin{figure}[!]
  \centering
  \begin{overpic}[height=5cm]{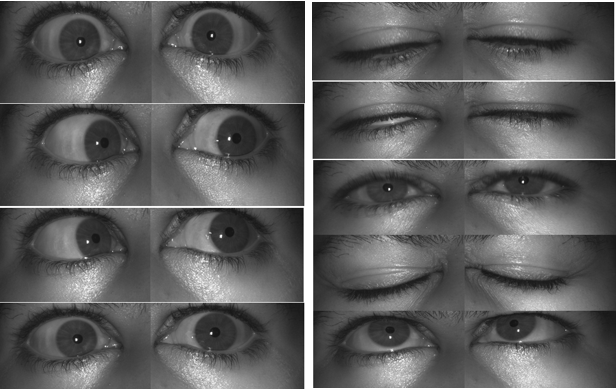}
  \end{overpic}
  \caption{ Exemplary samples of captured images in Step 1 (left image) and Step 2 (right image) }
 \label{fig:procedure} 
\end{figure}

The procedure for image capturing is in two steps:
\begin{itemize}
\item \emph{Step 1-Straight Gaze Capture: } In this step, the volunteer was asked to hold the sensor and gaze into its lenses while moving the eyes in small ranges to the left, right, up, or down for approximately 3 minutes. During this time, images were periodically captured automatically at constant intervals. The eye movement ensures that the iris position and orientation are highly diversified across the captured images. Samples of the captured images in this step are shown in  \figref{fig:procedure} (left).
 \item \emph{Step 2- Open-Close Capture: }  In this step, the volunteer was asked to open and close the eyes while gazing into the device for approximately 3 minutes. The iris images were automatically captured within this period. The essence of the opening and closing of the eyes is to ensure that the images contain irises of various sizes and degrees of occlusion, from fully closed eyes to half open and fully open eyes. This procedure will ultimately improve image diversity. Exemplary samples are shown in \figref{fig:procedure} (right).
\end{itemize}
All the iris images were acquired using this two steps procedure. As such, for every subject, there were two sets of images captured with a maximum of 84 images per set which means that up to 168 images were captured and saved for each subject.

\subsection{Database cleaning and construction}
\begin{figure}[!]
  \centering
  \begin{overpic}[width=0.7\linewidth]{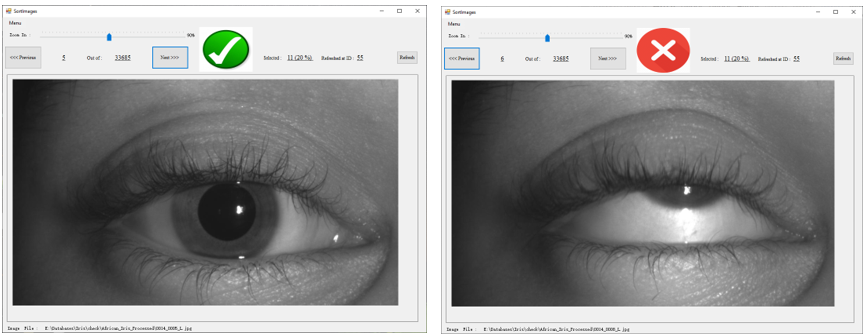}
  \end{overpic}
  \caption{ Selection tool used for manual processing of the images. }
 \label{fig:selection_tool}\vspace{-3pt}
\end{figure}

\begin{figure}[!]
  \centering
  \begin{overpic}[height=4cm]{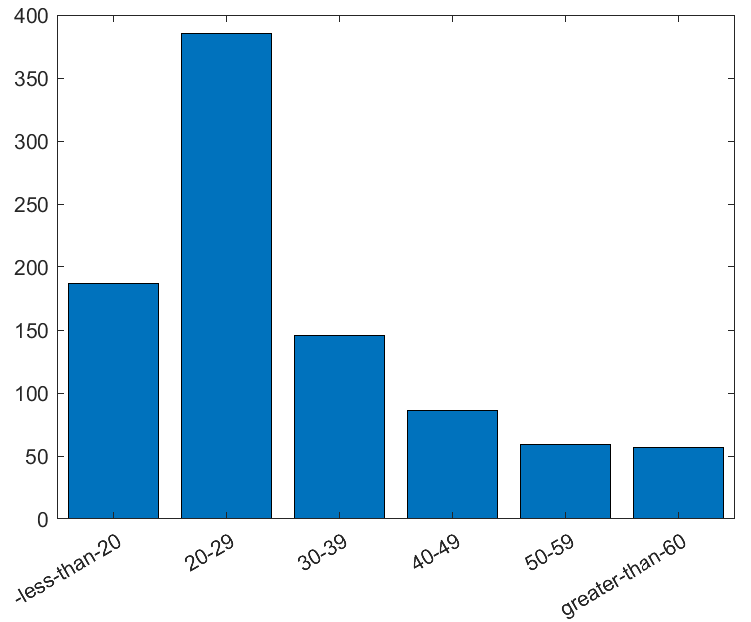}
  \end{overpic}
  \caption{ Age distribution of the generated database. }
 \label{fig:age_distribution}\vspace{-3pt}
\end{figure}

\begin{figure*}[t]
  \centering
  \begin{overpic}[width=\linewidth]{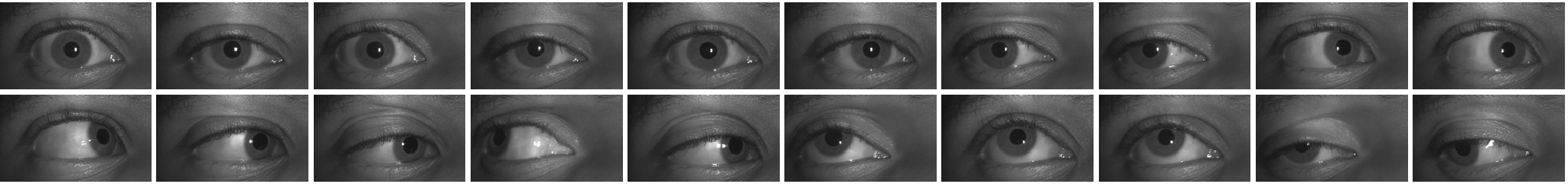}
  \end{overpic}
  \vspace{3pt}
  \begin{overpic}[width=\linewidth]{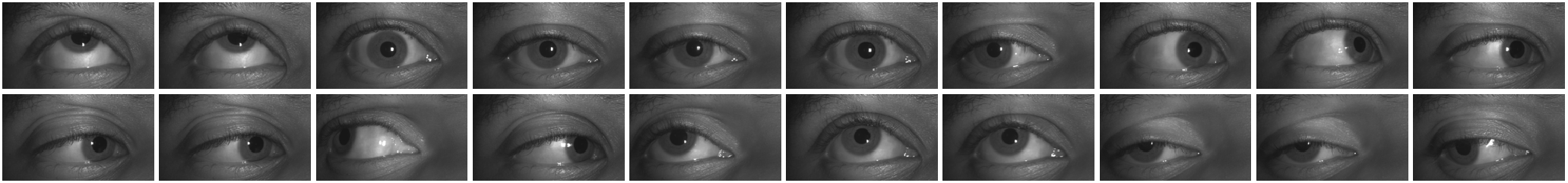}
  \end{overpic}
  \caption{ Samples of one subject left iris (upper two rows) and (b) right iris (lower two rows) }
 \label{fig:sample_images}\vspace{-3pt}
\end{figure*}

After the capturing sessions, the stored images were processed. The processing steps comprise automatic image selection to discard images with duplicate content, followed by manual image selection that guarantees the selected images' reliability. In the automatic selection, all the images from a single eye of one subject were organized into a matrix, with each column representing an image. The correlation coefficients of this matrix were then computed based on the technique proposed in \cite{teukolsky1992numerical}. The values of the computed coefficients range from 0 (no correlation) to 1(full correlation). Based on these coefficients, all pairs of images with coefficient values greater than the threshold $T$ are considered duplicates, and one of them must be discarded. A threshold of $T=0.82$ was heuristically chosen after careful adjustments with multiple values. The images of the subject's eyes were then processed, and duplicates were removed using this procedure. The remaining images were then manually processed. The manual selection involves one-by-one human inspection of each image to identify damaged images and those images with no irises captured in them. A GUI selection tool shown in \figref{fig:selection_tool} has been specially developed for this task, allowing efficient image selection. The tool provides the means of using just the mouse left click to select an image and right-click to discard an image. The selection decision is recorded on each mouse click, and the next image in the list is immediately loaded and displayed. This makes inspection very fast and can lead to the processing of up to 55 images per minute. All the images were inspected using this tool, and the unselected images were discarded. The remaining images constitute the generated dataset. Some sample images of one subject from the generated dataset are shown in \figref{fig:sample_images}, the summary of the database is presented in \tabref{tab:database_summary} and the database subjects' age distribution is shown in \figref{fig:age_distribution}.

\begin{table}[]
\centering
\renewcommand{\arraystretch}{1.1}
\setlength\tabcolsep{3pt}
\caption{ Summary of the generated database }
\label{tab:database_summary}
\begin{tabular}{p{0.5\linewidth}|p{0.2\linewidth}}
\hline
  Number of Subjects  & 1023 \\
  Male-Female Ratio  & 56\%-44\% \\
  Number of Classes  & 2046 \\
  Number of Images  & 28717 \\

  Resolution of Images & $1088 \times 640$\\ \hline
\end{tabular}
\end{table}

\subsection{Database labelling }
\begin{figure*}[t]
  \centering
  \begin{overpic}[width=0.98\linewidth]{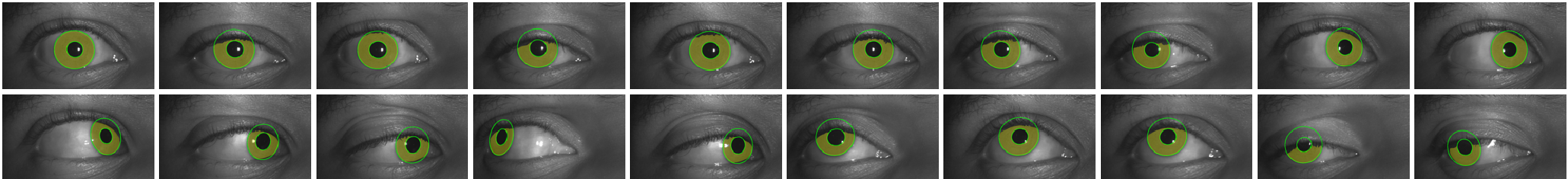}
  \end{overpic}
  \vspace{3pt}
  \begin{overpic}[width=0.98\linewidth]{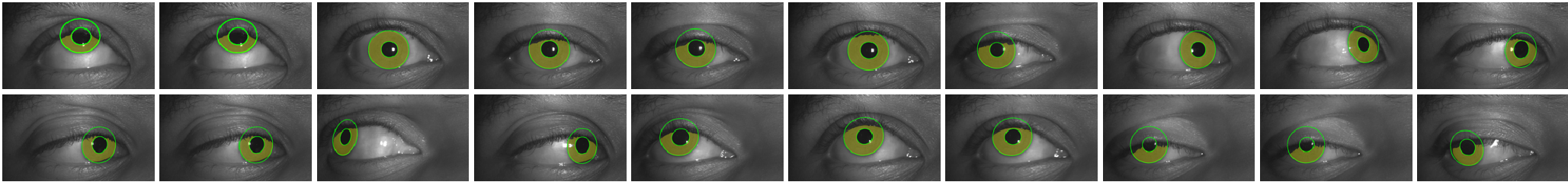}
  \end{overpic}
  \caption{ Labelled iris mask, inner and outer circle superimposed on respective samples in \figref{fig:sample_images} for (a) left iris (upper two rows) and (b) right iris (lower two rows) }
 \label{fig:sample_leblled_images}\vspace{-3pt}
\end{figure*}

After the database has been generated and organized, each iris image needs to be appropriately labelled to be effectively utilized for iris recognition tasks. The labelling process comprises two tasks: (1) pixel-wise labelling of the iris region to generate an iris segmentation mask; and (2) drawing of circular contours around the iris and pupil regions to yield the iris inner and outer cycle parameters. These tasks are tedious and delicate as they entail carefully segregating the boundaries around the iris region from the incursion of eyelids and eyelashes. We adopt a semiautomatic labelling approach. First, 1010 images were carefully selected and manually labelled as discussed in \cite{Wang2020Towards}. These images were then used to train two CNN-based models performing segmentation and localization. The trained models were then used to label all the remaining images in the database automatically. The GUI tool presented in \figref{fig:selection_tool} was again used to inspect the labelled images manually, and all the observed labelling errors were manually relabelled. Sample images of the labelled database are shown in \figref{fig:sample_leblled_images}.

\begin{figure}[!]
  \centering
  \begin{overpic}[width=\linewidth]{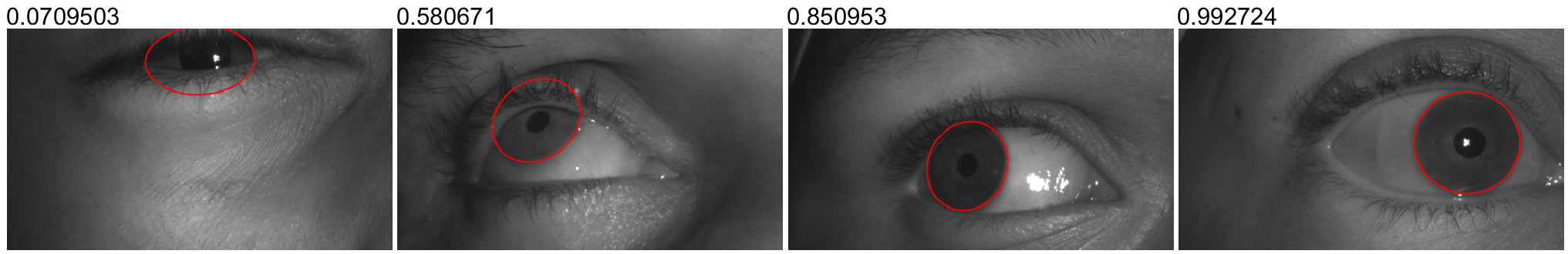}
  \end{overpic}
  \caption{ Some sample images with their corresponding openness scores }
 \label{fig:sample_completness_scores}\vspace{-3pt}
\end{figure}

\begin{figure}[]
  \centering
  \begin{overpic}[width=0.7\linewidth]{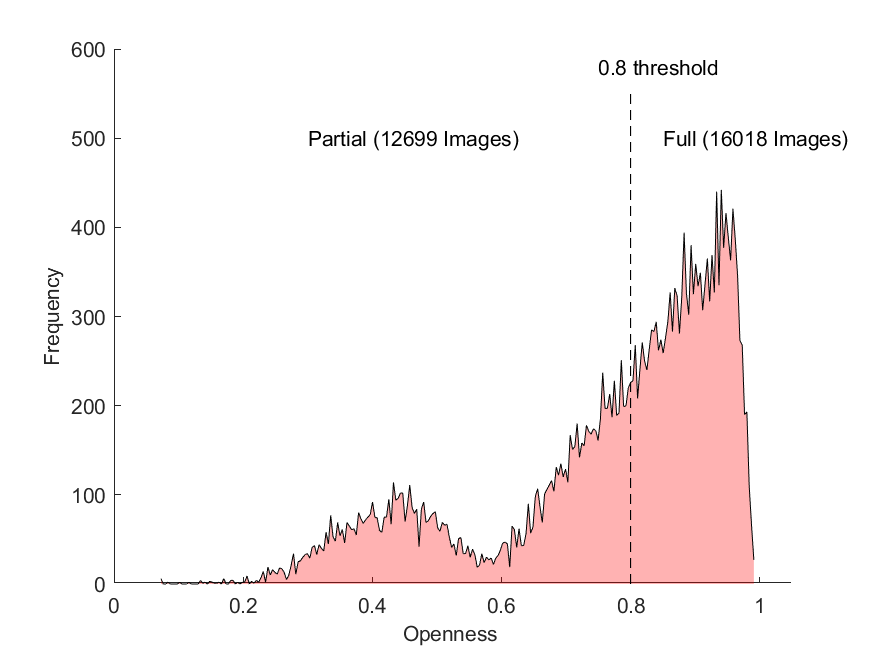}
  \end{overpic}
  \caption{ Distribution of the openness scores across the images of the proposed database showing the cut-off threshold }
 \label{fig:completness_distribution}\vspace{-3pt}
\end{figure}

After the labelling,  each image has a corresponding segmentation mask and iris inner and outer cycle parameters. Additionally,  an openness score is added to provide more contextual information on the database composition. It is based on the labelling information of each iris image. The openness score is an empirical measure of iris visibility. Since the iris images of the same subject were deliberately captured at various degrees of visibility/sizes, as discussed in the previous section, incorporating this information will be important in analyzing methods evaluated with the proposed database. The openness score is computed as the ratio of the number of pixels covered by the mask to the number of pixels covered by the iris outer cycle. As such, an image of a wide-open eye will have an iris with an openness score close to 1, and a partially closed eye will have an iris with an openness score much less than 1. Mathematically, the openness score is the complement of the pixelwise XOR operation between the segmentation mask $M$, and the iris outer boundary masks $O$ (filled cycle of the outer boundary). It can be written as:

\begin{align}\label{eq:evaluation_1}
  Openness &= 1 - \frac{1}{n}\sum_{i}\sum_{j}M\otimes O
\end{align}

where $n$ corresponds to the number of pixels covered by the iris outer boundary mask in the image.
In \figref{fig:sample_completness_scores}, the openness scores for some sample images are shown. As expected, it can be observed that the partially closed eye (\figref{fig:sample_completness_scores}(a)) has a very low visibility of about 0.07. In contrast, wide-open eyes have very high visibility of about 0.99. The distribution of the openness scores of the images in the generated dataset is shown in \figref{fig:completness_distribution}. Based on these values, a cut-off threshold of 0.8 was chosen to divide the dataset into iris images with partial (P) openness (score below 0.8) and iris images with full (F) openness (above 0.8). As such, the attribute of either P or F is added to the labelling information of each image in the database.

\subsection{ Application oriented database protocols }
\begin{figure}[]
  \centering
  \begin{overpic}[width=0.6\linewidth]{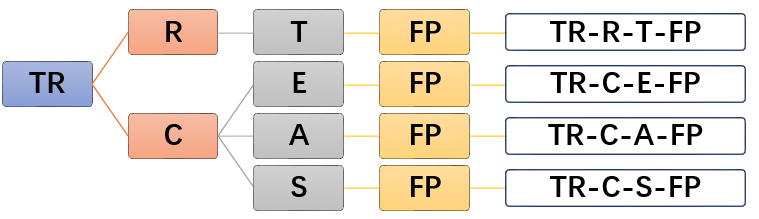}
  \end{overpic}
  \caption{ Database Training Protocol Tree }
 \label{fig:training_protocol_map}\vspace{-3pt}
\end{figure}

Inspired by \cite{muhammad2021casia}, a group of protocols is proposed, ensuring consistent results in adopting the proposed database. Unlike the strategy of defining protocols as a list of images by the existing proposed databases, this paper proposes application-oriented protocols that attempt to be closely related to the real-world application of iris biometrics. First, the database subjects are divided equally into training and testing partitions with no overlap of subjects between the two sets. The division is performed so that the subject's attributes are adequately represented across the two partitions. The protocols for the testing partition are shown in the protocol tree of \figref{fig:testing_protocol_map} and those of the training partition are shwon in  \figref{fig:training_protocol_map}. 
These protocols are fundamentally based on two factors: (1) database application for either recognition (R) or classification (C) tasks and (2) openness of either full (F) or partial (P) as shown in \figref{fig:testing_protocol_map} and \figref{fig:training_protocol_map}.  

\subsubsection{ Recognition protocols}
Each recognition protocol has a gallery setting and a query set emulating the real-life scenario of enrolled subjects (gallery set) and search subjects (query set) in an iris recognition application. The three main recognition tasks covered by the proposed protocols are: (1) close-set identification (I); (2) open set identification (O); and (3) verification (V) as shown in \figref{fig:testing_protocol_map} and \figref{fig:training_protocol_map}.

\begin{itemize}
  \item The close-set identification protocols contain one image per subject in the gallery set and multiple images per subject in the query set. A subject must have images in both sets. As such, ranking-based evaluation metrics can be used to evaluate images in this group of protocols.
  \item The open-set identification protocol is similar to the close-set, except that a subject may not have images in the gallery set. This is analogous to the identification of subjects who have not been previously enrolled. A combination of ranking-based evaluation metrics with a predefined threshold can be used to evaluate these groups of protocols.
  \item For the verification protocols, all images of subjects are included in both the gallery and query set such that mated and non-mated pair combinations can be generated from them. Many of these pair combinations with highly similar pairs can lead to redundant results. Therefore, to reduce the redundancy, the pairs were generated by maximizing the intraclass distance and minimizing the inter-class distances. This can be achieved using the absolute difference of the openness score of the images as the measure of pair similarity to choose only the highly dissimilar pairs of the mated pairs and discard the highly dissimilar pairs of the non-mated pairs.
\end{itemize}

\subsubsection{ Classification protocols}
The classification groups of protocols are based on the three available subject attributes in the database : (1) ethnicity (E); (2) age (A); and (3) sex (S) as shown in \figref{fig:testing_protocol_map} and \figref{fig:training_protocol_map}. Unlike in recognition, the classification protocol has only one set of images that can be used to evaluate the respective classification models.
For ethnicity classification, each of the six major ethnicities with a relatively high number of subjects was considered an individual class, while all the remaining ethnicities were categorized as others. This is necessary as some ethnicities have very few subjects which can lead to class unbalancing.
All the distinct subject age values are considered separate classes for age classification. This will provide the flexibility of either performing age group classification or age regression.
For the sex classification, the classes are the binary male or female classes. All the subjects can be classified to either one of these values.

\subsubsection{ Iris openness in protocols}
In addition to the database application considered for protocols generation, the iris image openness attribute described in the previous section has also been considered. The intuition here is to objectively evaluate algorithms' sensitivity to the iris's visibility in the African irises' context. The pair order permutations of PP, FF, FP, and PF have been considered for recognition protocols that require image pair comparisons. This corresponds to both images in the pair having partial openness attributes (PP), full (FF), or the query image full and gallery image partial (FP), and vice versa in PF. For the classification protocols, the image openness attribute is directly considered as the classification task typically involves the analysis of a single image.

\subsubsection{ Ideal subset protocols }
\begin{figure}[]
  \centering
  \begin{overpic}[height=3cm ]{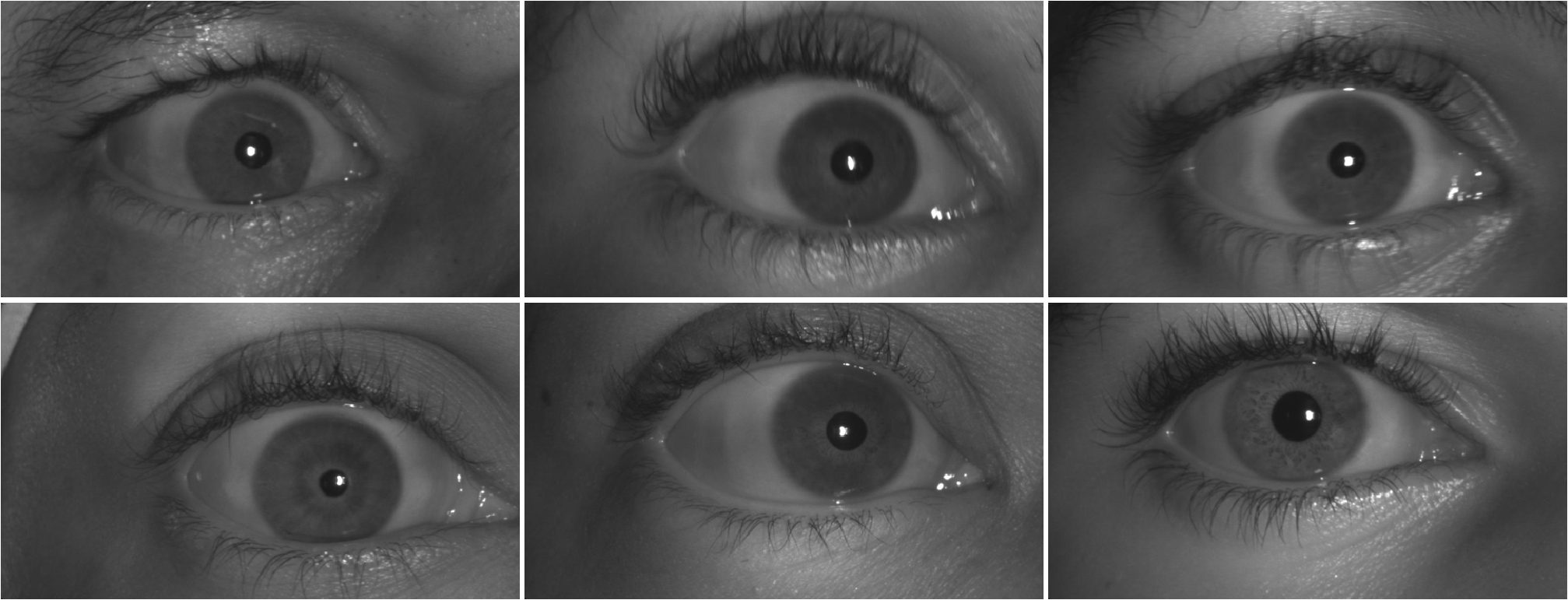}
  \end{overpic}
  \caption{ Sample images of the selected ideal set }
 \label{fig:ideal_Set}\vspace{-3pt}
\end{figure}

To provide a simplified set of images that can be quickly evaluated by algorithms, an ideal subset of images is generated by selecting images from the testing set with complete visibility ( openness score greater than 0.89) and a reasonably fixed front elevation (iris orientation $<=0.4$ radians ). This set is adopted in both the recognition and identification protocols. Sample images of the ideal set are shown in \figref{fig:ideal_Set}. 

\subsubsection{ Final protocols }
\begin{figure}[]
  \centering
  \begin{overpic}[width=0.6\linewidth]{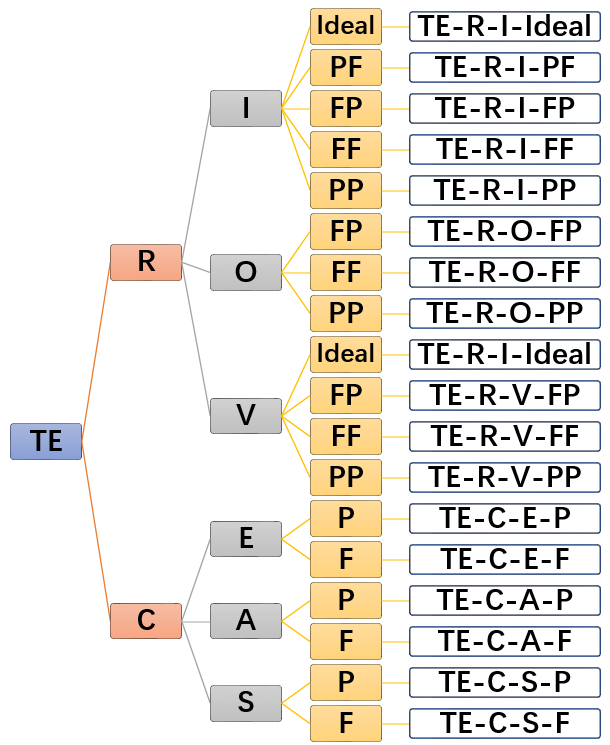}
  \end{overpic}
  \caption{ Database Testing Protocol Tree }
 \label{fig:testing_protocol_map}\vspace{-3pt}
\end{figure}

After considering all the stated factors, the complete set of protocols is shown in \figref{fig:testing_protocol_map} as the tree leaves, which comprises a total of 18 sets of protocols for the testing and 4 sets of protocols for the training set. Prospective users can choose one more of these protocols to adopt the proposed database appropriately. 

\section{Experiments and Baseline Performances}
\label{sec:baseline_performance}

\subsection{ Baseline algorithms }
To provide benchmark evaluations of the proposed database, state-of-the-art (SOTA) open-source algorithms  have been evaluated as baseline algorithms for the proposed database. The results will be presented in the next section. The baseline algorithms for the localization and segmentation task include the popular USIT \cite{USIT3} open source iris software package algorithms of USIT-Wahet, USIT-Caht, USIT-Cahtvis, and USIT-Ifpp. Other algorithms are MASK-RCNN \cite{ahmad2018unconstrained}, TVMiris\cite{zhao2015accurate}, OSIRIS\cite{othman2016osiris} and 2 of the top 3 best participants algorithms of the NIR-ISL 2021 iris recognition challenge\footnote{\href{https://sites.google.com/view/nir-isl2021/home}{https://sites.google.com/view/nir-isl2021/home}} tagged as NIR-ISL2021030902 and NIR-ISL2021041402\textunderscore2 \cite{wang2021nir}.
  The baseline algorithms for the iris recognition task also include some of the USIT implementation algorithms such as USIT-Lg; USIT-Cg; USIT-Qsw; USIT-Ko; USIT-Cr; USIT-Cb; and USIT-Dct. Other algorithms include UniNet \cite{zhao2017towards}, Graphnet\cite{Ren2020DynamicGR}, LightCNN\cite{Wu2015ALC}, Maxout\cite{Zhang2018DeepFF}, and Afnet\cite{Ren2019AlignmentFA}.

Many researchers have widely adopted all these algorithms, and their implementations are readily available. As such, it will be particularly appropriate to present the baseline performance of the algorithms on the proposed database.

\subsection{Iris segmentation and localization }
The baseline results of the localization and segmentation algorithms were evaluated against the ground truth of the constructed dataset. The evaluation criteria used for localization are the Hausdorff distance and Dice metrics. Hausdorff distance measures the shape similarity between the ground truth and predicted inner/outer boundary shapes, while Dice measures the overlap between these shapes. For the segmentation task, E1 and E2 error rates are used \cite{wang2021nir}. E1 is the average proportion of corresponding disagreeing pixels over all the images. It is calculated by the pixelwise XOR operation between the predicted segmentation mask and its ground truth \cite{wang2021nir}. At the same time, E2 is the average of false-positive and false-negative rates. The generated results are in two categories: (1) results of USIT and Mask-RCNN algorithms; and (2)results of NIR-ISL 2021 algorithms.
\begin{itemize}
  \item Category 1: the results of the USIT and Mask-RCNN algorithms are shown in \tabref{tab:pre_processing_results_1}. All the USIT-based algorithms perform poorly, while the deep learning-based Mask-RCNN algorithm has a relatively better error rate.
  \item Category 2: The algorithms from the NIR-ISL 2021 competition each have three models trained on three different datasets. These three models are tagged as Mod-1, Mod-2, and Mod-3, and the results are shown in \tabref{tab:pre_processing_results_2}. It can be observed that some of these models perform poorly while some exhibit relatively better performance.
  
\end{itemize}

\begin{table}[]
\centering
\renewcommand{\arraystretch}{1.1}
\setlength\tabcolsep{3pt}
\caption{ Iris Segmentation and Localization Results-Category 1 }
\label{tab:pre_processing_results_1}

\begin{tabular}{c|c|c }
\hline
\multirow{2}{*}{Algorithms} & \multicolumn{2}{c}{ Localization } \\ \cline{2-3}
  & Hausdorff  & Dice  \\ \hline


 USIT-Wahet\cite{USIT3}	 & 0.220	 &0.293\\ \hline
 TVMiris\cite{zhao2015accurate}	 &0.033	 &0.842\\ \hline
 OSIRIS\cite{othman2016osiris}	 &\textbf{0.027}	 &\textbf{0.848}\\ \hline


 \\ \cline{2-3}
& \multicolumn{2}{c}{ Segmentation } \\ \cline{2-3}
   & E1(\%) & E2(\%) \\ \hline


MASK-RCNN\cite{ahmad2018unconstrained}	             &2.289	 &1.145\\ \hline
TVMiris	                 &2.148	 &1.074\\ \hline
OSIRIS\cite{othman2016osiris}	                 &\textbf{1.687}	 &\textbf{0.844}\\ \hline
USIT-Caht\cite{USIT3}	 &5.442	 &2.722\\ \hline
USIT-Cahtvis\cite{USIT3} &5.566	 &2.783\\ \hline
USIT-Ifpp\cite{USIT3}	 &5.429	 &2.715\\ \hline

\end{tabular}
\end{table}

\begin{table}[]
\centering
\renewcommand{\arraystretch}{1.1}
\setlength\tabcolsep{3pt}
\caption{ Iris Segmentation and Localization Results-Category 2 }
\label{tab:pre_processing_results_2}

\begin{tabular}{c|c|c|c||c|c|c}
\hline
\multirow{3}{*}{Algorithms} & \multicolumn{6}{|c}{ Localization } \\ \cline{2-7}
  & \multicolumn{3}{c||}{ Hausdorff } & \multicolumn{3}{c}{  Dice }\\ \cline{2-7}
   &  Mod-1	&	Mod-2	&	Mod-3	&	Mod-1	&	Mod-2	&	Mod-3	\\ \hline


NIR-ISL2021030902	                 &\textbf{0.007}	 &\textbf{0.011}	 &\textbf{0.007}	 &\textbf{0.971}	 &\textbf{0.952}	 &\textbf{0.971}\\ \hline
NIR-ISL2021041402\textunderscore2	 &0.023	 &0.018	 &0.019	 &0.853	 &0.920	 &0.912\\ \hline


 \\ \cline{2-7}
& \multicolumn{6}{|c}{ Segmentation } \\ \cline{2-7}
   & \multicolumn{3}{c||}{ E1(\%) } & \multicolumn{3}{c}{ E2(\%) }\\ \cline{2-7}
&  Mod-1	&	Mod-2	&	Mod-3	&	Mod-1	&	Mod-2	&	Mod-3	\\ \hline


NIR-ISL2021030902	                 &\textbf{0.299}	 &\textbf{0.390}	 &\textbf{0.299}	 &\textbf{0.149}	 &\textbf{0.195}	 &\textbf{0.149}\\ \hline
NIR-ISL2021041402\textunderscore2	 &2.960	 &0.909	 &1.629	 &1.480	 &0.455	 &0.815\\ \hline

\end{tabular}
\end{table}

\subsection{Iris recognition}
Before extracting the iris representation from the images used for iris recognition evaluations, the images of the generated database were segmented and localized using the ground truth labelLing data from the database. Each image was first segmented and normalized to a rectangular image of size $64 \times 512$ using Daugman's rubber sheet model \cite{Daugman2009How}.

\subsubsection{ Verification results }
\begin{figure}
\vspace{2pt}
  \centering
  \begin{overpic}[width=\linewidth ]{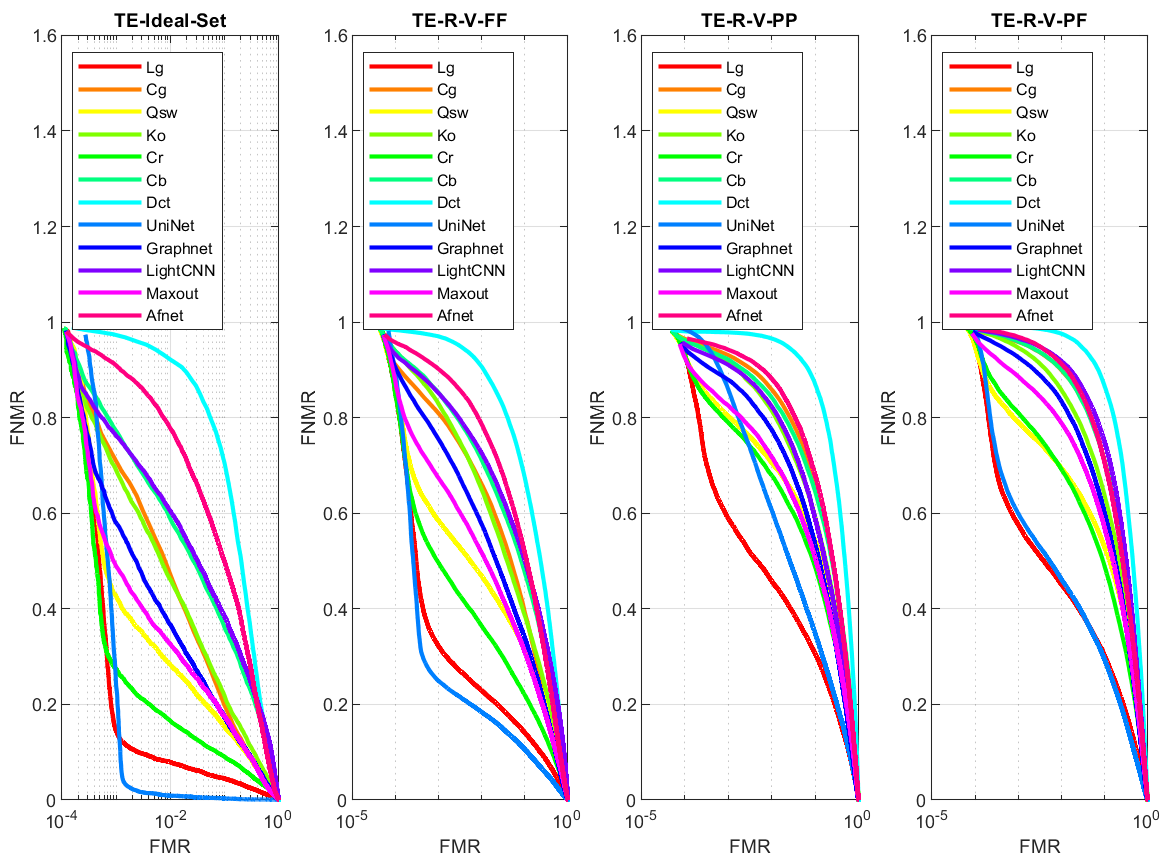}
  \end{overpic}
  \caption{ DET curves for protocols: Ideal-Set, TE-R-V-PF, TE-R-V-FF, and TE-R-V-PP  }
 \label{fig:iris_verification_DET}\vspace{-3pt}
\end{figure}

The results for the verification of the TE-R-V-xx group of protocols are presented in \tabref{tab:verification_TE_R_V} and their corresponding detection error tradeoff (DET) curves are shown in \figref{fig:iris_verification_DET}. It can be observed that the baseline algorithm performances are generally weak across all the protocols. Almost all the algorithms fail at 0.01\% FMR, which is a standing operating point for real-time application. The results of the algorithms are relatively better on the TE-R-V-FF protocol and worse on the TE-R-V-PF protocol. This can be attributed to the fact that the pairs in TE-R-V-FF have images with similar openness scores and hence similar iris visibility between the FF pairs, while in TE-R-V-PF, it is the opposite case. These results demonstrate the inherent challenges in off-the-shelf algorithms on niche use cases such as African irises. Hence, the proposed database's availability is needed so that researchers can fine-tune their algorithms.

\begin{sidewaystable}
\centering
\renewcommand{\arraystretch}{1.1}
\setlength\tabcolsep{3pt}
\caption{ TE-R-V- protocols results: FNMR(\%) at various values of FMR }
\label{tab:verification_TE_R_V}
 
\begin{tabular}{|c|c|c|c|c|c|c|c|c|c|c|c|c|c|c|c|c|}
\hline
\multirow{2}{*}{Algorithm} & \multicolumn{4}{c|}{Ideal Set} & \multicolumn{4}{c|}{TE-R-V-FF }                                      & \multicolumn{4}{c|}{TE-R-V-PP} & \multicolumn{4}{c|}{TE-R-V-PF }    
\\ \cline{2-17}
& EER     & 10\%     & 1\%     & 0.01\% &  EER     & 10\%     & 1\%     & 0.01\%
                           & EER     & 10\%     & 1\%     & 0.01\%
                           & EER     & 10\%     & 1\%     & 0.01\%
                            \\ \hline


USIT-Lg    & 5.21 & 4.42 & 7.88 & 14.35 & 12.87 & 14.07 & 22.88 & 32.13 & 22.60 & 30.63 & 45.78 & 58.82 & 22.32 & 30.23 & 45.13 & 57.61 \\
USIT-Cg    & 14.67 & 18.55 & 45.50 & 68.85 & 23.98 & 39.16 & 65.52 & 80.01 & 37.95 & 65.82 & 85.31 & 93.04 & 38.75 & 69.71 & 89.67 & 95.48 \\
USIT-Qsw   & 13.59 & 15.33 & 27.96 & 42.79 & 23.24 & 31.21 & 46.84 & 58.44 & 35.51 & 54.16 & 70.58 & 80.27 & 34.79 & 52.94 & 69.44 & 79.11 \\
USIT-Ko    & 16.56 & 21.06 & 45.85 & 68.82 & 25.69 & 38.12 & 64.86 & 82.34 & 35.36 & 58.66 & 81.46 & 91.22 & 37.16 & 63.69 & 85.34 & 94.49 \\
USIT-Cr    & 9.21 & 8.94 & 16.45 & 27.37 & 17.66 & 21.91 & 36.36 & 49.51 & 31.37 & 48.67 & 67.92 & 78.94 & 31.17 & 48.92 & 69.20 & 80.92 \\
USIT-Cb    & 24.70 & 35.39 & 58.12 & 77.04 & 30.13 & 47.25 & 71.00 & 84.94 & 38.78 & 64.55 & 84.02 & 92.12 & 40.84 & 70.35 & 90.08 & 95.78 \\
USIT-Dct   & 32.27 & 70.29 & 91.37 & 96.74 & 37.10 & 67.14 & 90.50 & 96.54 & 49.04 & 86.85 & 96.54 & 97.83 & 47.48 & 86.56 & 97.50 & 98.92 \\ \hline 
UniNet     & \textbf{0.90} & \textbf{0.27} & \textbf{0.85} & \textbf{24.05} & \textbf{10.37} & \textbf{10.57} & \textbf{18.35} & \textbf{24.86} & \textbf{23.87} & \textbf{34.93} & \textbf{60.71} & \textbf{85.91} & \textbf{21.23} & \textbf{29.44} & \textbf{46.20} & \textbf{59.94} \\
Graphnet   & 14.44 & 17.24 & 36.66 & 58.00 & 23.02 & 33.83 & 57.68 & 75.44 & 31.40 & 53.39 & 77.40 & 87.92 & 34.31 & 60.12 & 83.46 & 92.62 \\
LightCNN   & 26.51 & 38.31 & 60.44 & 76.17 & 32.28 & 50.02 & 72.56 & 85.10 & 34.71 & 58.88 & 82.33 & 91.04 & 41.30 & 75.00 & 91.83 & 96.59 \\
Maxout     & 14.48 & 17.48 & 32.61 & 48.90 & 21.85 & 30.94 & 51.72 & 67.32 & 31.70 & 50.67 & 71.81 & 82.54 & 34.31 & 57.05 & 78.89 & 88.61 \\
Afnet      & 28.83 & 49.81 & 77.78 & 90.89 & 29.92 & 51.65 & 78.60 & 90.13 & 38.74 & 68.04 & 88.11 & 94.20 & 38.86 & 70.76 & 91.28 & 96.91 \\ \hline       


\hline
\end{tabular}
\end{sidewaystable}

\subsubsection{Identification results }
\begin{figure}
\vspace{2pt}
  \centering
  \begin{overpic}[width=\linewidth ]{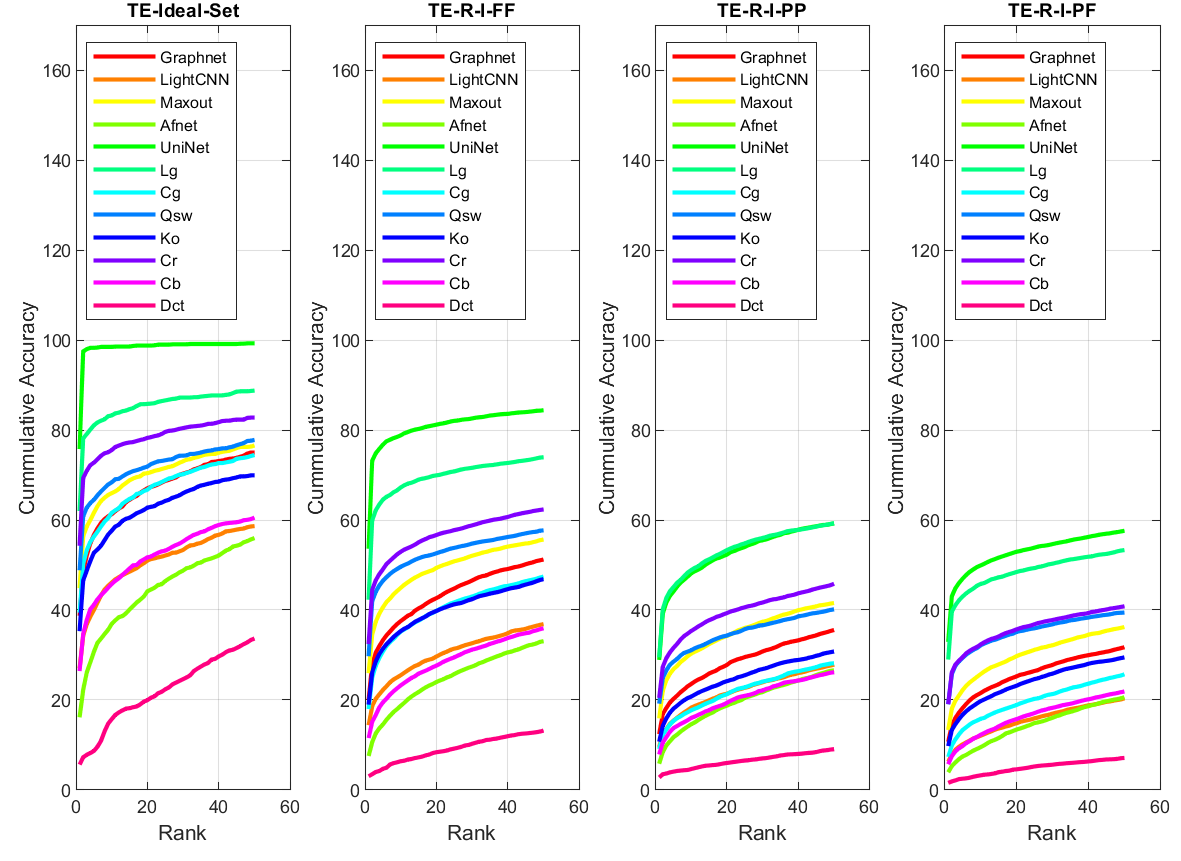}
  \put (14,-1) { (a) }
  \put (37,-1) { (b) }
  \put (62,-1) { (c) }
  \put (87,-1) { (c) }
  \end{overpic}
  \caption{ Identification results: CMC curves for the protocols: (a) TE-R-I-FP (b) TE-R-I-PF  (c) TE-R-I-FF and (d) TE-R-I-PP  }
 \label{fig:iris_CMC}\vspace{-3pt}
\end{figure}

The results for the identification protocols of TE-R-I- are presented in \tabref{tab:open_identification_TE_R_I} and their corresponding cumulative matching characteristics (CMC) curves are shown in \figref{fig:iris_CMC}. The poor performances of the algorithms across the protocols coincide with the outcome of the verification results. The results are slightly better for the TE-R-I-FF protocol than for TE-R-I-FP and TE-R-I-PF due to similar visibility on TE-R-I-FF pair comparisons. However, it is interesting to observe that the results of TE-R-I-PF are generally relatively better than those of TE-R-I-FP. This gives an interesting insight: probing with a partially occluded African iris (P openness) against a less occluded iris (F openness) yields better results. The presented identification results further demonstrate the value of the proposed database as a complementary database of Africans by the iris biometric research community.

\begin{sidewaystable}
\centering
\renewcommand{\arraystretch}{1.1}
\setlength\tabcolsep{3pt}
\caption{ Identification results in (\%) for protocols TE-R-I-}
\label{tab:open_identification_TE_R_I}
\begin{tabular}{|c|c|c|c||c|c|c||c|c|c||c|c|c||c|c|c|}
\hline

\multirow{2}{*}{Algorithm} & \multicolumn{3}{c|}{Ideal Set (\%) } & \multicolumn{3}{c|}{TE-R-I-FF (\%) } & \multicolumn{3}{c|}{TE-R-I-PP (\%)} &  \multicolumn{3}{c|}{TE-R-I-FP (\%) } &  \multicolumn{3}{c|}{TE-R-I-PF (\%) }  
\\ \cline{2-16}
    & R-1 & R-5 & R-10 & R-1 & R-5 & R-10 & R-1 & R-5 & R-10  & R-1 & R-5 & R-10
     & R-1 & R-5 & R-10
    \\ \hline

USIT-Lg   & 61.99 & 81.07 & 83.26 & 42.27 & 64.53 & 67.24 & 29.45 & 45.08 & 48.89 & 25.45 & 35.65 & 38.31 & 28.97 & 41.74 & 44.42 \\
USIT-Cg   & 39.41 & 56.31 & 61.60 & 17.96 & 30.74 & 35.23 & 9.07 & 15.34 & 17.70 & 5.84 & 10.16 & 12.75 & 7.31 & 12.88 & 15.28 \\
USIT-Qsw  & 48.83 & 64.49 & 68.38 & 29.71 & 46.49 & 49.62 & 19.16 & 28.51 & 31.04 & 16.60 & 24.27 & 26.80 & 19.03 & 28.62 & 31.29 \\
USIT-Ko   & 35.28 & 52.73 & 57.63 & 18.94 & 31.40 & 35.30 & 10.69 & 18.06 & 20.74 & 7.29 & 11.81 & 14.38 & 9.72 & 16.27 & 19.19 \\
USIT-Cr   & 54.28 & 72.74 & 75.55 & 32.41 & 49.04 & 52.94 & 20.44 & 31.57 & 35.35 & 15.96 & 23.11 & 26.19 & 19.05 & 28.66 & 31.37 \\
USIT-Cb   & 26.40 & 41.12 & 45.79 & 11.52 & 19.47 & 23.23 & 7.86 & 13.74 & 16.00 & 4.30 & 7.86 & 9.70 & 5.69 & 9.53 & 11.94 \\
USIT-Dct  & 5.61 & 8.64 & 15.73 & 3.02 & 4.69 & 6.37 & 2.76 & 4.04 & 4.63 & 1.33 & 1.73 & 2.25 & 1.55 & 2.48 & 3.21 \\ \hline 
UniNet    & \textbf{75.77} & \textbf{98.28} & \textbf{98.50} & \textbf{53.61} & \textbf{76.76} & \textbf{78.82} & \textbf{28.88} & \textbf{44.09} & \textbf{48.17} & \textbf{31.13} & \textbf{44.88} & \textbf{47.78} & \textbf{32.89} & \textbf{46.95} & \textbf{49.90} \\
Graphnet  & 38.71 & 56.70 & 61.29 & 19.94 & 33.28 & 37.53 & 12.39 & 20.22 & 23.56 & 8.11 & 12.77 & 15.26 & 10.66 & 17.63 & 20.91 \\
LightCNN  & 27.02 & 39.95 & 46.11 & 14.37 & 22.22 & 25.49 & 9.52 & 15.40 & 18.31 & 4.90 & 7.62 & 9.50 & 6.26 & 9.89 & 11.75 \\
Maxout    & 44.86 & 61.60 & 65.97 & 25.93 & 40.59 & 44.91 & 15.89 & 26.85 & 30.30 & 11.10 & 17.55 & 20.12 & 13.31 & 21.54 & 24.99 \\
Afnet     & 16.12 & 30.69 & 36.92 & 7.52 & 14.54 & 18.69 & 5.80 & 11.69 & 14.60 & 3.38 & 6.85 & 8.80 & 3.90 & 7.33 & 9.34 \\ \hline 

\end{tabular}
\end{sidewaystable}

\subsection{ Genuine and impostor distribution  }

\begin{figure}
\vspace{2pt}
  \centering
  \begin{overpic}[width=0.7\linewidth ]{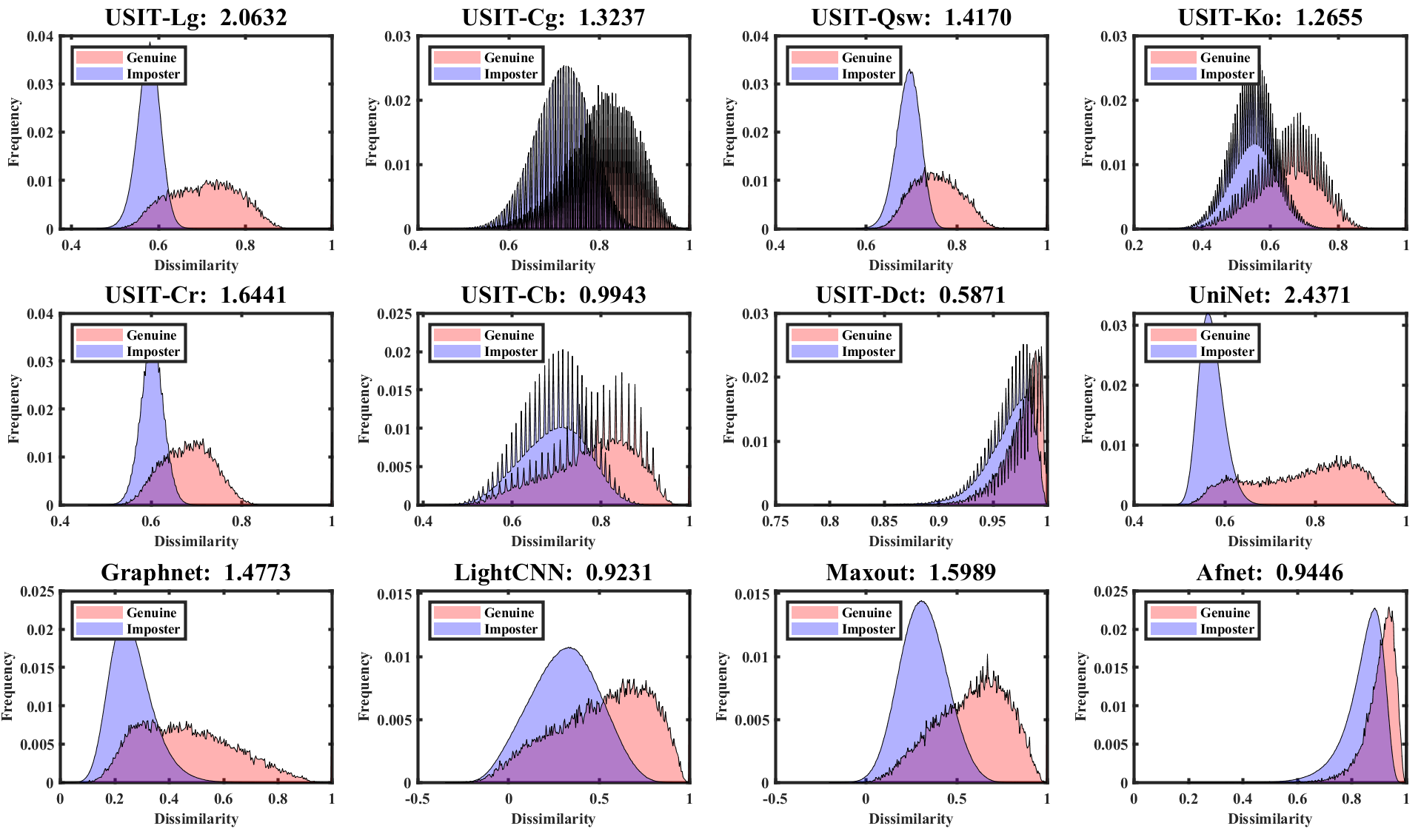}
  \end{overpic}
  \caption{ Genuine-impostor pairs distribution for TE-R-V-FF  }
 \label{fig:TE_R_V_FF_genuine_impostor_distribution}\vspace{-3pt}
\end{figure}

\begin{figure}
\vspace{2pt}
  \centering
  \begin{overpic}[width=0.7\linewidth ]{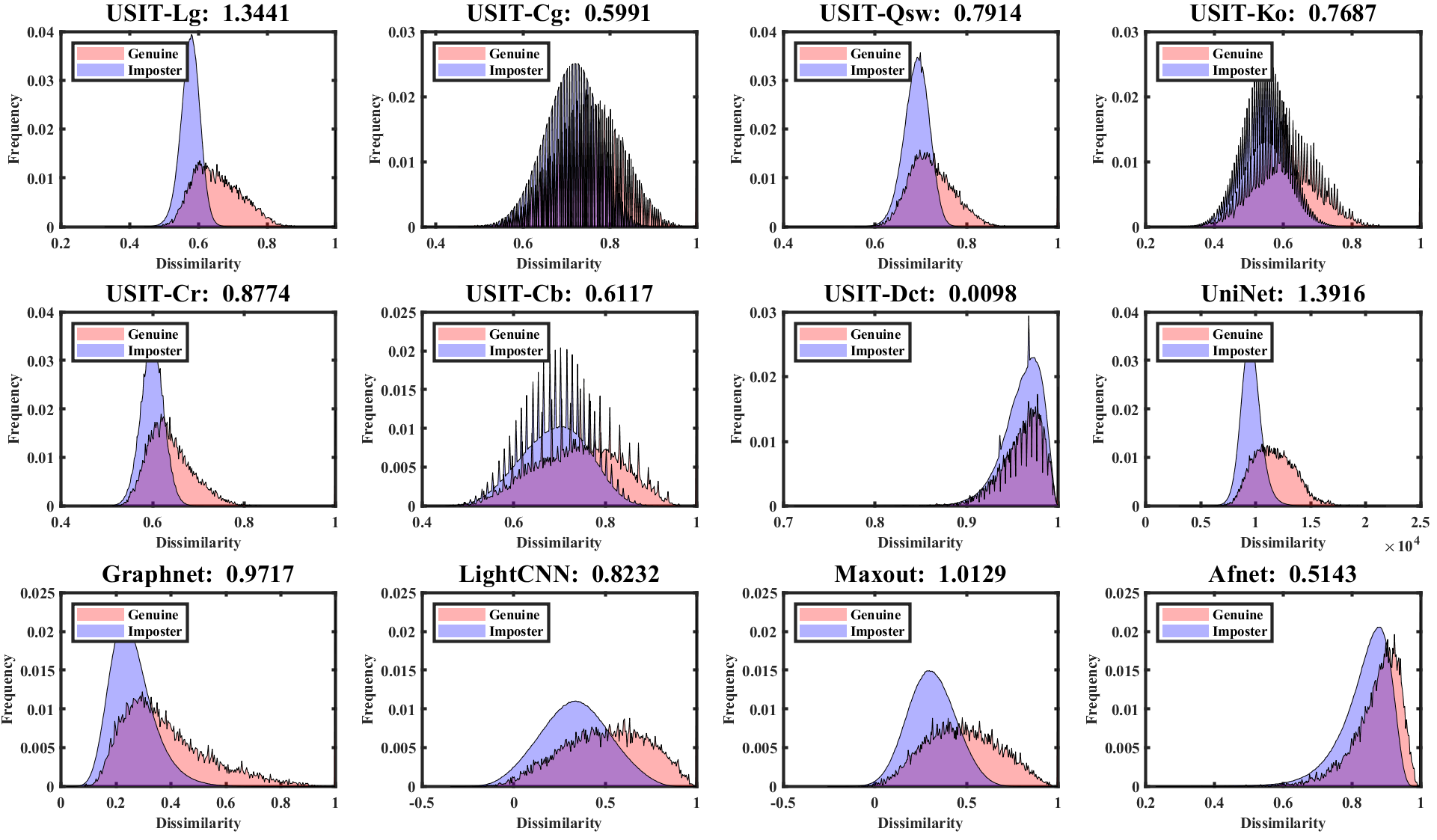}
  \end{overpic}
  \caption{ Genuine-impostor pairs distribution for TE-R-V-PP  }
 \label{fig:TE_R_V_PP_genuine_impostor_distribution}\vspace{-3pt}
\end{figure}

To analyse the performance of the baseline algorithms,  the genuine-imposter distribution of the algorithms with their respective Decidability Index (DI)\cite{daugman1998recognizing} on the protocols TE-R-V-PF, TE-R-V-FF and TE-R-V-PP are shown in \figref{fig:TE_R_V_PP_genuine_impostor_distribution}, \figref{fig:TE_R_V_FF_genuine_impostor_distribution} and \figref{fig:TE_R_V_PP_genuine_impostor_distribution} respectively. This distribution can be used to analyze the iris representations from these algorithms. It can be observed that, as evident from the previous results, the features of the TE-R-V-FF protocol extracted by the algorithms show much better genuine-imposter pair separation with a DI of up to 1.8269 on some algorithms. The lowest DIs are all from the TE-R-V-PF protocols. In general, all the protocols have bad imposter-genuine separation by all the algorithms.

\begin{figure}
\vspace{2pt}
  \centering
  \begin{overpic}[width=0.7\linewidth ]{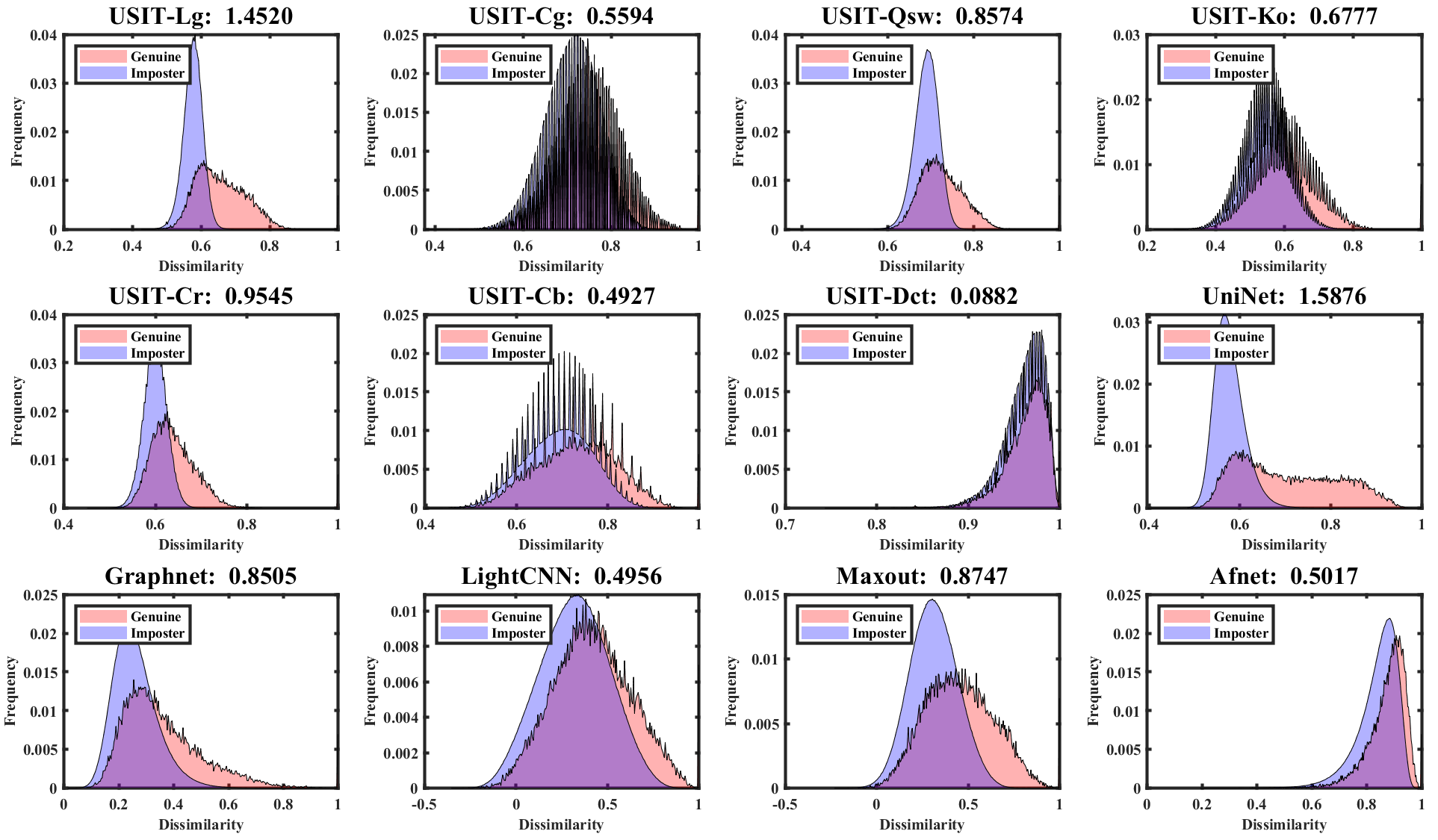}
  \end{overpic}
  \caption{ Genuine-impostor pairs distribution for TE-R-V-PF }
 \label{fig:TE_R_V_PF_genuine_impostor_distribution}\vspace{-3pt}
\end{figure}

\subsection{ Demographic analysis  }
To further analyse the effect of the database demographic attributes on the baseline results, the age and sex demographics performance of these algorithms is presented in \tabref{tab:identification_demog_sex} and \tabref{tab:identification_demog_age} respectively. For the sex demographics, it can be observed that the result for the female cohorts is consistently better than that of the males across all the algorithms and protocols. For some algorithms, the deviation is up to 100\%. For the age demographics, it can be observed that the baseline results of the age group \emph{$<20$ years} is consistently lower than all others across the protocols and algorithms, while that on the $>60$  age groups is consistently better. It is important to note that fewer samples of the$>60$ age group are in the database. 

\begin{sidewaystable}
\centering
\renewcommand{\arraystretch}{1.1}
\setlength\tabcolsep{3pt}
\caption{ Sex demographic analysis in (\%) for the protocols TE-R-I-}
\label{tab:identification_demog_sex}
\begin{tabular}{|c|c|c|c|c||c|c|c|c||c|c|c|c||c|c|c|c|}
\hline

\multirow{2}{*}{Algorithm} & \multicolumn{4}{c|}{TE-Ideal-Set (\%)} &  \multicolumn{4}{c|}{TE-R-I-FF (\%)} & \multicolumn{4}{c|}{TE-R-I-PP (\%) } &  \multicolumn{4}{c|}{TE-R-I-PF (\%)}
\\ \cline{2-17}
&  \multicolumn{2}{c|}{Rank 1} &  \multicolumn{2}{c|}{Rank 5} & \multicolumn{2}{c|}{Rank 1} & \multicolumn{2}{c|}{Rank 5} &  \multicolumn{2}{c|}{Rank 1} &  \multicolumn{2}{c|}{Rank 5} & \multicolumn{2}{c|}{Rank 1} & \multicolumn{2}{c|}{Rank 5}
\\ \cline{2-17}
    & M & FM & M & FM & M & FM & M & FM & M & FM & M & FM & M & FM & M & FM \\ \hline

USIT-Lg & 64.09 & 83.38 & 70.65 & 82.26 & 45.63 & 64.98 & 50.15 & 69.69 & 31.53 & 48.11 & 33.97 & 48.80 & 31.66 & 41.80 & 37.86 & 48.73 \\
USIT-Cg & 42.21 & 60.37 & 43.87 & 61.29 & 19.90 & 32.33 & 24.17 & 38.76 & 10.06 & 16.73 & 11.96 & 18.75 & 7.94 & 13.23 & 10.65 & 18.90 \\
USIT-Qsw & 50.67 & 67.60 & 53.23 & 67.42 & 32.68 & 47.24 & 37.12 & 53.26 & 20.45 & 31.21 & 22.17 & 30.27 & 20.94 & 29.20 & 26.53 & 34.28 \\
USIT-Ko & 37.46 & 56.35 & 43.55 & 55.48 & 19.36 & 30.42 & 28.42 & 42.88 & 11.34 & 19.37 & 14.35 & 22.77 & 10.92 & 17.47 & 12.76 & 21.58 \\
USIT-Cr & 56.86 & 76.26 & 60.65 & 74.52 & 33.73 & 49.08 & 42.54 & 58.43 & 22.33 & 34.37 & 23.97 & 34.40 & 20.58 & 28.96 & 26.68 & 34.82 \\
USIT-Cb & 29.21 & 44.07 & 33.87 & 49.03 & 12.04 & 20.41 & 16.86 & 27.20 & 8.73 & 15.78 & 9.46 & 15.76 & 6.27 & 10.61 & 7.03 & 13.59 \\
USIT-Dct & 5.99 & 9.80 & 9.03 & 23.87 & 3.50 & 5.75 & 3.15 & 7.52 & 3.19 & 4.70 & 2.88 & 4.35 & 1.58 & 2.50 & 2.58 & 4.20 \\ \hline 
UniNet & 81.66 & 98.64 & 85.82 & 97.52 & 56.13 & 78.12 & 63.04 & 79.23 & 32.49 & 47.82 & 33.42 & 47.16 & 35.34 & 49.77 & 37.76 & 48.23 \\
Graphnet & 36.47 & 57.43 & 44.89 & 64.72 & 20.91 & 36.11 & 22.22 & 37.96 & 12.47 & 20.96 & 15.77 & 26.29 & 10.15 & 17.93 & 15.05 & 24.60 \\
LightCNN & 26.43 & 41.17 & 30.90 & 48.43 & 14.95 & 24.28 & 16.27 & 25.00 & 9.74 & 16.33 & 11.76 & 19.83 & 4.98 & 9.13 & 10.85 & 16.64 \\
Maxout & 43.96 & 62.90 & 48.23 & 69.10 & 28.05 & 43.94 & 26.15 & 43.92 & 16.59 & 27.49 & 18.84 & 33.16 & 12.92 & 22.01 & 19.43 & 29.78 \\
Afnet & 17.79 & 34.56 & 20.46 & 37.58 & 8.75 & 17.56 & 9.39 & 19.40 & 6.38 & 12.94 & 7.81 & 16.50 & 3.43 & 7.31 & 7.27 & 12.63 \\ \hline 

\hline

\end{tabular}
\end{sidewaystable}

\begin{sidewaystable}
\centering
\renewcommand{\arraystretch}{1.1}
\setlength\tabcolsep{3pt}
\caption{ Age Demographic Analysis in (\%) for the protocols TE-R-I-}
\label{tab:identification_demog_age}
\begin{tabular}{|c|c|c|c|c|c|c||c|c|c|c|c|c|}
\hline

\multirow{2}{*}{Algorithm} & \multicolumn{12}{c|}{Rank 1 Identification (\%)} \\ \cline{2-13}
& \multicolumn{6}{c|}{TE-R-I-FP} & \multicolumn{6}{c|}{TE-R-I-FF} \\ \cline{2-13}
& less 20 & 20-29 & 30-39 & 40-49 & 50-59 & greater 60 & less 20 & 20-29 & 30-39 & 40-49 & 50-59 & greater 60 \\ \hline

USIT-Lg & 61.77 & 48.78 & 59.12 & 59.38 & 54.35 & 56.28 & 48.50 & 37.12 & 42.45 & 37.48 & 40.33 & 41.85 \\
USIT-Cg & 33.39 & 21.50 & 31.75 & 27.50 & 25.23 & 32.79 & 13.17 & 11.03 & 10.71 & 12.58 & 12.42 & 25.79 \\
USIT-Qsw & 46.99 & 36.74 & 43.96 & 42.81 & 38.74 & 38.14 & 33.65 & 26.28 & 29.06 & 23.41 & 29.53 & 29.93 \\
USIT-Ko & 31.98 & 22.87 & 30.81 & 29.37 & 25.53 & 36.74 & 16.90 & 13.19 & 15.99 & 12.58 & 19.14 & 22.14 \\
USIT-Cr & 46.21 & 37.38 & 48.93 & 45.78 & 43.24 & 52.09 & 31.24 & 24.61 & 30.36 & 22.06 & 32.99 & 34.55 \\
USIT-Cb & 19.39 & 14.78 & 19.43 & 23.59 & 18.02 & 24.42 & 9.66 & 8.30 & 10.55 & 10.01 & 13.65 & 13.38 \\
USIT-Dct & 7.90 & 3.34 & 8.18 & 2.03 & 4.50 & 12.09 & 2.56 & 1.83 & 2.84 & 3.38 & 2.85 & 6.57 \\ \hline 
UniNet & \textbf{62.55} & \textbf{59.60} & \textbf{73.88} & \textbf{66.55} & \textbf{69.12} & \textbf{79.04} & \textbf{35.81} & \textbf{39.82} & \textbf{46.91} & \textbf{31.17} & \textbf{52.98} & \textbf{53.16} \\
Graphnet & 25.16 & 19.99 & 26.10 & 28.35 & 47.32 & 30.82 & 13.15 & 11.17 & 15.76 & 13.75 & 33.07 & 26.49 \\
LightCNN & 15.38 & 14.62 & 19.90 & 22.33 & 33.75 & 24.21 & 10.44 & 5.95 & 10.26 & 10.47 & 11.20 & 15.41 \\
Maxout & 28.14 & 27.03 & 36.26 & 33.40 & 51.42 & 32.39 & 15.70 & 14.24 & 19.23 & 19.15 & 37.33 & 30.54 \\
Afnet & 10.68 & 8.58 & 14.64 & 17.09 & 32.18 & 19.18 & 6.84 & 4.55 & 7.99 & 7.36 & 13.87 & 10.54 \\ \hline 

\hline

\end{tabular}
\end{sidewaystable}

\subsection{ Comparison with other databases  }
\begin{figure}
\vspace{2pt}
  \centering
  \begin{overpic}[width=0.7\linewidth ]{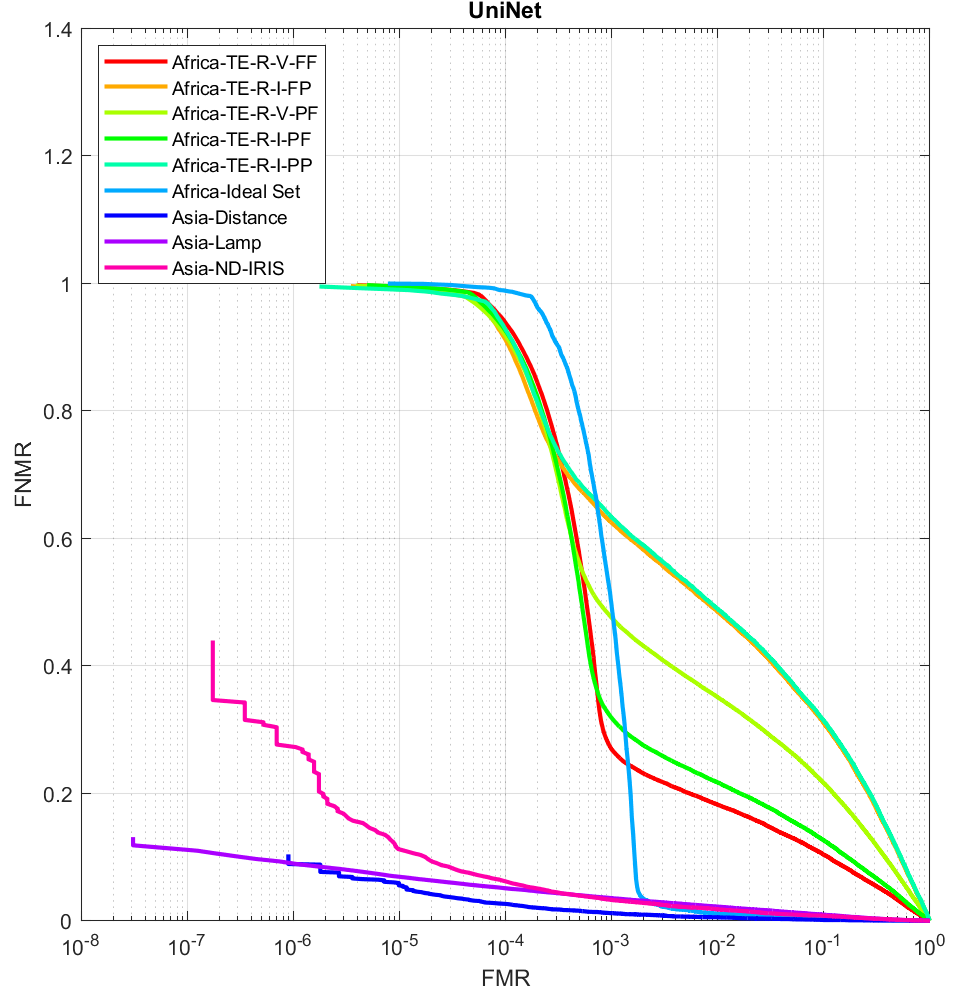}
  \end{overpic}
  \caption{ DET Curve for Database Comparisions }
 \label{fig:crossDB_DET_curve}\vspace{-3pt}
\end{figure}

Observing how the baseline algorithms perform poorly on the proposed database, the same algorithms have been evaluated on the subsets of some of the publicly available datasets. The selected databases are ND-IRIS-0405 (ND-Iris)  \cite{bowyer2016nd}, CASIA IrisV3-Distance (Distance) \cite{casia_v3}, and CASIA IrisV3-Lamp (Lamp) \cite{casia_v3}. For fair comparisons, the same localization, segmentation, normalization, and feature extraction strategy are used across all the databases. The Uninet model results were adopted for this comparison. The results of the verification DET curve are shown in \figref{fig:crossDB_DET_curve}  and the identification results are shown in  \tabref{tab:databases_comparison_identification}. 
Despite performing relatively better on most of the other databases, the model performs poorly on the proposed dataset protocols. This further demonstrates the racial biases of these algorithms. Even though the model has better performance on the Ideal set protocol, at higher FNMR, the performance is poor, similar to the other proposed database protocols.

%



\begin{table}[]
\centering
\renewcommand{\arraystretch}{1.1}
\setlength\tabcolsep{3pt}
\caption{  Database comparisons of baseline identification results  }
\label{tab:databases_comparison_identification}
\begin{tabular}{|c|c|c|c|c|}
\hline

Algorithm &  EER(\%) & Rank-1  & Rank-5 & Rank-10  \\ \hline

Africa-TE-R-V-FF & 10.25 & 54.96 & 93.40 & 94.67 \\
Africa-TE-R-I-FP & 22.16 & 45.12 & 75.01 & 78.18 \\
Africa-TE-R-V-PF & 17.24 & 52.39 & 87.63 & 90.03 \\
Africa-TE-R-I-PF & 11.77 & 52.65 & 89.38 & 91.20 \\
Africa-TE-R-I-PP & 22.40 & 45.07 & 74.58 & 77.78 \\
Africa-Ideal Set & 1.07 & 58.03 & 90.66 & 91.43 \\  \hline 
Asia-Distance    & \textbf{0.60} & \textbf{99.67} & \textbf{99.73} & \textbf{99.73} \\
Asia-Lamp        & 1.85 & 99.73 & 99.80 & 99.83 \\
Asia-ND-IRIS     & 1.55 & 99.26 & 99.56 & 99.68 \\

\hline 
\end{tabular}
\end{table}

\section{Conclusions and Future Work}
\label{sec::conclusion}

This paper presents a large-scale African database that can be used to mitigate racial bias in African iris biometrics. The database contains 1,023 subjects with 2046 unique irises and 28,717 images. A database application protocol has been presented to provide the variability and repeatability of results across algorithms. The baseline performances of some SOTA algorithms have also been presented, proving the existence of iris racial bias in some of these algorithms.
In future work, a more comprehensive analysis of African iris biometrics will be conducted using the proposed CASIA-Iris-Africa, and algorithms will be developed by utilizing the database to address the racial bias problem of African iris biometrics.

\section*{Acknowledgements}

This work was supported in part by the National Natural Science Foundation of China under Grants: 62176025, 62071468, 62006225, in part by Strategic Priority Research Program of the Chinese Academy of Sciences under Grant XDA27040700, and the CAS-TWAS President’s Fellowship for International Doctoral Students

\begin{figure}[h]%
\centering
\includegraphics[width=0.3\textwidth]{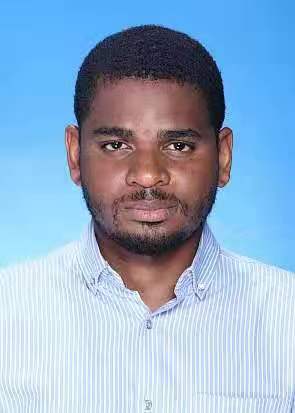}
\end{figure}

\noindent{\bf Jawad~Muhammad} received the B.E.~degree in Computer Engineering from Bayero university, Kano, Nigeria, in 2010, and the M.S. degree in Computer Engineering from Selcuk University, Konya, Turkey, in 2014. He is currently pursuing the Ph.D. degree with the School of Artificial Intelligence, University of Chinese Academy of Sciences, China, and the Center for Research on Intelligent Perception and Computing, National Laboratory of Pattern Recognition, Institute of Automation, Chinese Academy of Sciences. His current research interests include biometrics, computer vision, and robot vision.
E-mail: jawad@cripac.ia.ac.cn
ORCID iD: 0000-0003-2547-5267

\begin{figure}[h!]%
\centering
\includegraphics[width=0.3\textwidth]{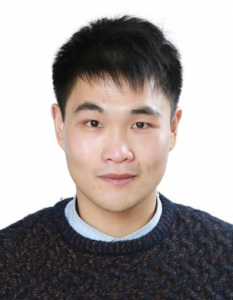}
\end{figure}

\noindent{\bf Yunlong Wang} received the B.E.~and Ph.D.~degrees from the Department of Automation, University of Science and Technology of China. He is currently an associate professor with the Center for Research on Intelligent Perception and Computing, National Laboratory of Pattern Recognition, Institute of Automation, Chinese Academy of Sciences. His research focuses on pattern recognition, machine learning, light field photography, and biometrics. 
E-mail: yunlong.wang@cripac.ia.ac.cn
ORCID iD: 0000-0002-3535-308X

\begin{figure}[h!]%
\centering
\includegraphics[width=0.3\textwidth]{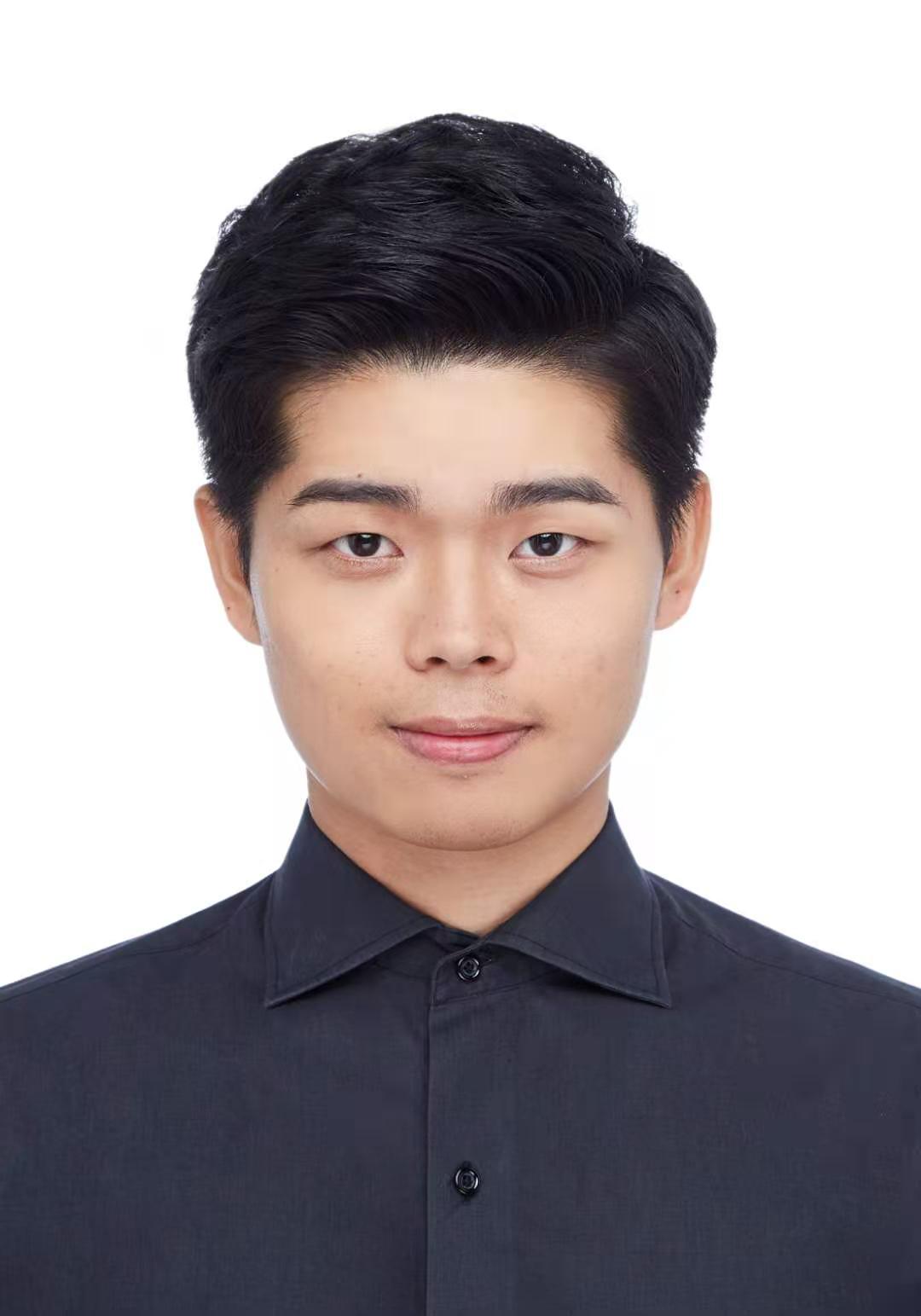}
\end{figure}

\noindent{\bf Junxing Hu}  is currently pursuing a Ph.D. degree with the School of Artificial Intelligence, University of Chinese Academy of Sciences, and the CRIPAC, NLPR, Institute of Automation, Chinese Academy of Sciences, China. He received a B.E. degree in software engineering from the Dalian University of Technology, in 2017, and an M.S. degree in computer science from the Institute of Software, Chinese Academy of Sciences, in 2020. His current research interests include biometrics, computer vision, and deep learning.
E-mail:junxing.hu@cripac.ia.ac.cn 
ORCID iD: 0000-0000-0000-0000

\begin{figure}[h!]%
\centering
\includegraphics[width=0.3\textwidth]{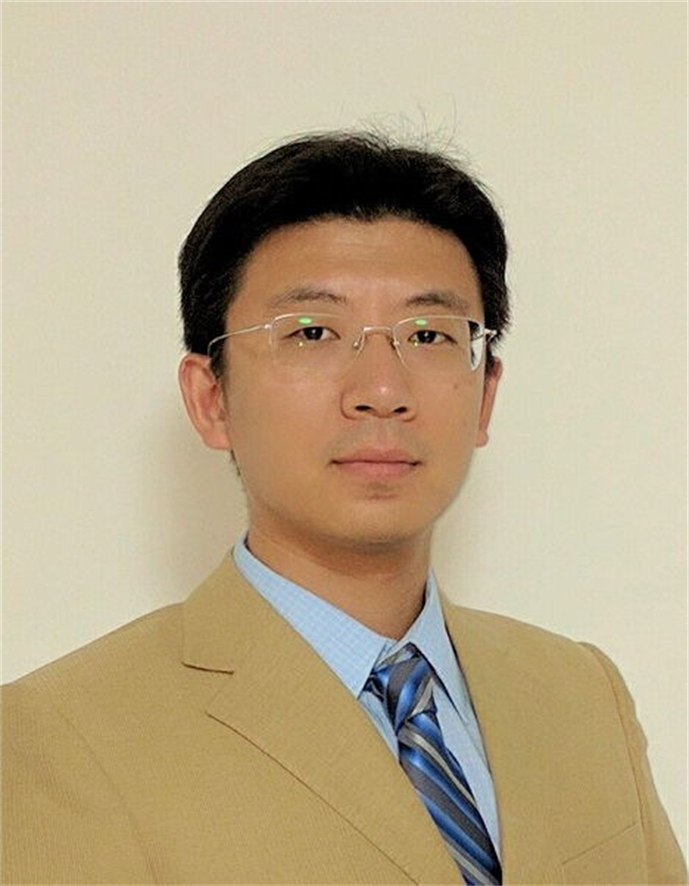}
\end{figure}

\noindent{\bf Kunbo Zhang} is currently an associate professor at CRIPAC, NLPR, CASIA, China. He received the B.E degree in Automation from Beijing Institute of Technology in 2006, and M.Sc and Ph.D. degrees in Mechanical Engineering from State University of New York at Stony Brook, U.S., in 2008 and 2011, respectively. Between 2011 and 2016, he worked as machine vision R\&D engineer of Advanced Manufacturing Engineering group in Nexteer Automotive, Michigan, U.S. His current research interests focus on computational photography, biometric imaging, machine vision, and intelligent systems.
E-mail: ; kunbo.zhang@ia.ac.cn
ORCID iD: 0000-0000-0000-0000

\begin{figure}[h!]%
\centering
\includegraphics[width=0.3\textwidth]{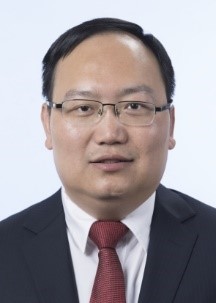}
\end{figure}

\noindent{\bf Zhenan Sun} (Senior Member, IEEE) received the B.E. degree in industrial automation from the Dalian University of Technology, China, in 1999, the M.S.
degree in system engineering from the Huazhong University of Science and Technology, China, in 2002, and the Ph.D. degree in pattern recognition and intelligent systems from the Institute of Automation, Chinese Academy of Sciences in 2006. He is currently a Professor with the Center for Research on Intelligent Perception and Computing, National Laboratory of Pattern Recognition, Institute of Automation, Chinese Academy of Sciences, China, and also with the School of Artificial Intelligence, University of Chinese Academy of Sciences, China. He has authored/coauthored over 200 technical papers. His research interests include biometrics, pattern recognition, and computer vision. He serves as an Associate Editor for the IEEE Transactions on Biometrics, Behaviour and Identity Science (T-BIOM). He is the Chair of Technical Committee on Biometrics, International Association for Pattern Recognition (IAPR) and an IAPR Fellow.
E-mail: znsun@nlpr.ia.ac.cn
ORCID iD: 0000-0000-0000-0000


\begin{thebibliography}{9}

\providecommand{\bibinfo}[2]{#2}
\providecommand{\BIBentrySTDinterwordspacing}{\spaceskip=0pt\relax}
\providecommand{\BIBentryALTinterwordstretchfactor}{4}
\providecommand{\BIBentryALTinterwordspacing}{\spaceskip=\fontdimen2\font plus
\BIBentryALTinterwordstretchfactor\fontdimen3\font minus
  \fontdimen4\font\relax}
\providecommand{\BIBforeignlanguage}[2]{{%
\expandafter\ifx\csname l@#1\endcsname\relax
\typeout{** WARNING: IEEEtran.bst: No hyphenation pattern has been}%
\typeout{** loaded for the language `#1'. Using the pattern for}%
\typeout{** the default language instead.}%
\else
\language=\csname l@#1\endcsname
\fi
#2}}
\providecommand{\BIBdecl}{\relax}
\BIBdecl

\bibitem{2008Image}
K.~W. Bowyer, K.~Hollingsworth, and P.~J. Flynn, ``Image understanding for iris
  biometrics: A survey,'' \emph{Computer Vision \& Image Understanding}, vol.
  110, no.~2, pp. 281--307, 2008.

\bibitem{omelina2021survey}
L.~Omelina, J.~Goga, J.~Pavlovicova, M.~Oravec, and B.~Jansen, ``A survey of
  iris datasets,'' \emph{Image and Vision Computing}, vol. 108, p. 104109,
  2021.

\bibitem{casia_v2}
\BIBentryALTinterwordspacing
 [Online]. Available:
  \url{http://biometrics.idealtest.org/dbDetailForUser.do?id=14#/datasetDetail/2}
\BIBentrySTDinterwordspacing

\bibitem{de2015mobile}
M.~De~Marsico, M.~Nappi, D.~Riccio, and H.~Wechsler, ``Mobile iris challenge
  evaluation (miche)-i, biometric iris dataset and protocols,'' \emph{Pattern
  Recognition Letters}, vol.~57, pp. 17--23, 2015.

\bibitem{VISOB_Dataset}
A.~Rattani, R.~Derakhshani, S.~K. Saripalle, and V.~Gottemukkula, ``Icip 2016
  competition on mobile ocular biometric recognition,'' in \emph{IEEE
  International Conference on Image Processing (ICIP) 2016, Challenge Session
  on Mobile Ocular Biometric Recognition}, 2016.

\bibitem{crossEyed7736915}
A.~Sequeira, L.~Chen, P.~Wild, J.~Ferryman, F.~Alonso-Fernandez, K.~B. Raja,
  R.~Raghavendra, C.~Busch, and J.~Bigun, ``Cross-eyed - cross-spectral
  iris/periocular recognition database and competition,'' in \emph{2016
  International Conference of the Biometrics Special Interest Group (BIOSIG)},
  2016, pp. 1--5.

\bibitem{Toward2616281}
P.~R. Nalla and A.~Kumar, ``Toward more accurate iris recognition using
  cross-spectral matching,'' \emph{IEEE transactions on image processing : a
  publication of the IEEE Signal Processing Society}, vol.~26, no.~1, pp.
  208--221, 2017.

\bibitem{WVU4100635}
J.~Zuo, N.~A. Schmid, and X.~Chen, ``On generation and analysis of synthetic
  iris images,'' \emph{IEEE Transactions on Information Forensics and
  Security}, vol.~2, no.~1, pp. 77--90, 2007.

\bibitem{casia_v4}
``Casia iris image database v3,''
  \url{http://biometrics.idealtest.org/dbDetailForUser.do?id=14#/datasetDetail/4},
  accessed: 2022-02-28.

\bibitem{UBIRISv2}
H.~Proenca, S.~Filipe, R.~Santos, J.~Oliveira, and L.~A. Alexandre, ``The
  ubiris.v2: A database of visible wavelength iris images captured on-the-move
  and at-a-distance,'' \emph{IEEE Transactions on Pattern Analysis and Machine
  Intelligence}, vol.~32, no.~8, pp. 1529--1535, 2010.

\bibitem{contact6776569}
D.~Yadav, N.~Kohli, J.~S. Doyle, R.~Singh, M.~Vatsa, and K.~W. Bowyer,
  ``Unraveling the effect of textured contact lenses on iris recognition,''
  \emph{IEEE Transactions on Information Forensics and Security}, vol.~9,
  no.~5, pp. 851--862, 2014.

\bibitem{liveness8272763}
D.~Yambay, B.~Becker, N.~Kohli, D.~Yadav, A.~Czajka, K.~W. Bowyer,
  S.~Schuckers, R.~Singh, M.~Vatsa, A.~Noore, D.~Gragnaniello, C.~Sansone,
  L.~Verdoliva, L.~He, Y.~Ru, H.~Li, N.~Liu, Z.~Sun, and T.~Tan, ``Livdet iris
  2017 - iris liveness detection competition 2017,'' in \emph{2017 IEEE
  International Joint Conference on Biometrics (IJCB)}, 2017, pp. 733--741.

\bibitem{aging6239214}
S.~P. Fenker and K.~W. Bowyer, ``Analysis of template aging in iris
  biometrics,'' in \emph{2012 IEEE Computer Society Conference on Computer
  Vision and Pattern Recognition Workshops}, 2012, pp. 45--51.

\bibitem{Patrick2019Face}
G.~Patrick, N.~Mei, and H.~Kayee, ``Face recognition vendor test (frvt) part 3:
  Demographic effects,'' National Institute of Standards and Technology, Report
  NISTIR 8280, 2019.

\bibitem{Buolamwini1}
J.~Buolamwini and T.~Gebru, ``Gender shades: Intersectional accuracy
  disparities in commercial gender classification,'' in \emph{Conference on
  fairness, accountability and transparency}, Conference Proceedings, pp.
  77--91.

\bibitem{cavazos2020accuracy}
J.~G. Cavazos, P.~J. Phillips, C.~D. Castillo, and A.~J. O'Toole, ``Accuracy
  comparison across face recognition algorithms: Where are we on measuring race
  bias?'' \emph{IEEE Transactions on Biometrics, Behavior, and Identity
  Science}, 2020.

\bibitem{bowyer2016nd}
K.~W. Bowyer and P.~J. Flynn, ``The nd-iris-0405 iris image dataset,''
  \emph{arXiv preprint arXiv:1606.04853}, 2016.

\bibitem{proencca2005ubiris}
H.~Proen{\c{c}}a and L.~A. Alexandre, ``Ubiris: A noisy iris image database,''
  in \emph{International Conference on Image Analysis and Processing}.\hskip
  1em plus 0.5em minus 0.4em\relax Springer, 2005, pp. 970--977.

\bibitem{KUMAR20101016}
\BIBentryALTinterwordspacing
A.~Kumar and A.~Passi, ``Comparison and combination of iris matchers for
  reliable personal authentication,'' \emph{Pattern Recognition}, vol.~43,
  no.~3, pp. 1016--1026, 2010. [Online]. Available:
  \url{https://www.sciencedirect.com/science/article/pii/S0031320309003343}
\BIBentrySTDinterwordspacing

\bibitem{IITDWebsite}
``Iit delhi iris database (version 1.0),''
  \url{https://www4.comp.polyu.edu.hk/~csajaykr/IITD/Database_Iris.htm},
  accessed: 2022-03-17.

\bibitem{phillips2008iris}
P.~J. Phillips, K.~W. Bowyer, P.~J. Flynn, X.~Liu, and W.~T. Scruggs, ``The
  iris challenge evaluation 2005,'' in \emph{2008 IEEE Second International
  Conference on Biometrics: Theory, Applications and Systems}.\hskip 1em plus
  0.5em minus 0.4em\relax IEEE, 2008, pp. 1--8.

\bibitem{mmu_db}
``Multimedia university version 2 iris database, 2010,''
  \url{http://pesona.mmu.edu.my/~ccteo/}, accessed: 2022-02-28.

\bibitem{monro2009university}
D.~Monro, S.~Rakshit, and D.~Zhang, ``University of bath, uk iris image
  database,'' 2009.

\bibitem{Ma02irisrecognition}
L.~Ma, Y.~Wang, and T.~Tan, ``Iris recognition based on multichannel gabor
  filtering,'' in \emph{Proceedings of the International Conference on Asian
  Conference on Computer Vision}, 2002, pp. 279--283.

\bibitem{casia_v3}
``Casia iris image database v3,''
  \url{http://biometrics.idealtest.org/dbDetailForUser.do?id=14#/datasetDetail/3},
  accessed: 2022-02-28.

\bibitem{Phillips26}
P.~J. Phillips, H.~Moon, S.~A. Rizvi, and P.~J. Rauss, ``The feret evaluation
  methodology for face-recognition algorithms,'' \emph{IEEE Transactions on
  pattern analysis and machine intelligence}, vol.~22, no.~10, pp. 1090--1104,
  2000.

\bibitem{li2021robust}
Y.-H. Li, W.~R. Putri, M.~S. Aslam, and C.-C. Chang, ``Robust iris segmentation
  algorithm in non-cooperative environments using interleaved residual u-net,''
  \emph{Sensors}, vol.~21, no.~4, p. 1434, 2021.

\bibitem{ahmad2018unconstrained}
S.~Ahmad and B.~Fuller, ``Unconstrained iris segmentation using convolutional
  neural networks,'' in \emph{Asian Conference on Computer Vision}.\hskip 1em
  plus 0.5em minus 0.4em\relax Springer, 2018, pp. 450--466.

\bibitem{Daugman2009How}
J.~Daugman, ``How iris recognition works,'' \emph{The Essential Guide to Image
  Processing (Second Edition)}, vol.~14, no.~1, pp. 715--739, 2009.

\bibitem{Tan2013Towards}
C.~W. Tan and A.~Kumar, ``Towards online iris and periocular recognition under
  relaxed imaging constraints,'' \emph{IEEE Transactions on Image Processing},
  vol.~22, no.~10, pp. 3751--3765, 2013.

\bibitem{Gangwar2016IrisSeg}
A.~Gangwar, A.~Joshi, A.~Singh, F.~Alonso-Fernandez, and J.~Bigun, ``Irisseg: A
  fast and robust iris segmentation framework for non-ideal iris images,'' in
  \emph{2016 International Conference on Biometrics (ICB)}, 2016, pp. 1--8.

\bibitem{Tan2012Unified}
C.~W. Tan and A.~Kumar, ``Unified framework for automated iris segmentation
  using distantly acquired face images,'' \emph{IEEE Trans Image Process},
  vol.~21, no.~9, pp. 4068--4079, 2012.

\bibitem{Radman2017Automated}
A.~Radman, N.~Zainal, and S.~A. Suandi, ``Automated segmentation of iris images
  acquired in an unconstrained environment using hog-svm and growcut,''
  \emph{Digital Signal Processing}, vol.~64, pp. 60--70, 2017.

\bibitem{Proena2010Iris}
H.~Proena, ``Iris recognition: On the segmentation of degraded images acquired
  in the visible wavelength,'' \emph{IEEE Trans Pattern Anal Mach Intell},
  vol.~32, no.~8, pp. 1502--1516, 2010.

\bibitem{Sardar2020Iris}
M.~Sardar, S.~Banerjee, and S.~Mitra, ``Iris segmentation using interactive
  deep learning,'' \emph{IEEE Access}, vol.~8, pp. 219\,322--219\,330, 2020.

\bibitem{Wang2020Towards}
C.~Wang, J.~Muhammad, Y.~Wang, Z.~He, and Z.~Sun, ``Towards complete and
  accurate iris segmentation using deep multi-task attention network for
  non-cooperative iris recognition,'' \emph{IEEE Transactions on Information
  Forensics and Security}, vol.~15, pp. 2944--2959, 2020.

\bibitem{fogel1989gabor}
I.~Fogel and D.~Sagi, ``Gabor filters as texture discriminator,''
  \emph{Biological cybernetics}, vol.~61, no.~2, pp. 103--113, 1989.

\bibitem{masek2003recognition}
L.~Masek \emph{et~al.}, ``Recognition of human iris patterns for biometric
  identification,'' Ph.D. dissertation, Citeseer, 2003.

\bibitem{sun2008ordinal}
Z.~Sun and T.~Tan, ``Ordinal measures for iris recognition,'' \emph{IEEE
  Transactions on pattern analysis and machine intelligence}, vol.~31, no.~12,
  pp. 2211--2226, 2008.

\bibitem{ahmed1974discrete}
N.~Ahmed, T.~Natarajan, and K.~R. Rao, ``Discrete cosine transform,''
  \emph{IEEE transactions on Computers}, vol. 100, no.~1, pp. 90--93, 1974.

\bibitem{monro2007dct}
D.~M. Monro, S.~Rakshit, and D.~Zhang, ``Dct-based iris recognition,''
  \emph{IEEE transactions on pattern analysis and machine intelligence},
  vol.~29, no.~4, pp. 586--595, 2007.

\bibitem{miyazawa2008effective}
K.~Miyazawa, K.~Ito, T.~Aoki, K.~Kobayashi, and H.~Nakajima, ``An effective
  approach for iris recognition using phase-based image matching,'' \emph{IEEE
  transactions on pattern analysis and machine intelligence}, vol.~30, no.~10,
  pp. 1741--1756, 2008.

\bibitem{sundararajan2001discrete}
D.~Sundararajan, \emph{The discrete Fourier transform: theory, algorithms and
  applications}.\hskip 1em plus 0.5em minus 0.4em\relax World Scientific, 2001.

\bibitem{gangwar2016deepirisnet}
A.~Gangwar and A.~Joshi, ``Deepirisnet: Deep iris representation with
  applications in iris recognition and cross-sensor iris recognition,'' in
  \emph{2016 IEEE international conference on image processing (ICIP)}.\hskip
  1em plus 0.5em minus 0.4em\relax IEEE, 2016, pp. 2301--2305.

\bibitem{zhao2017towards}
Z.~Zhao and A.~Kumar, ``Towards more accurate iris recognition using deeply
  learned spatially corresponding features,'' in \emph{Proceedings of the IEEE
  international conference on computer vision}, 2017, pp. 3809--3818.

\bibitem{zhang2018deep}
Q.~Zhang, H.~Li, Z.~Sun, and T.~Tan, ``Deep feature fusion for iris and
  periocular biometrics on mobile devices,'' \emph{IEEE Transactions on
  Information Forensics and Security}, vol.~13, no.~11, pp. 2897--2912, 2018.

\bibitem{teukolsky1992numerical}
W.~H. Press, B.~P. Flannery, S.~A. Teukolsky, and W.~T. Vetterling, ``Numeric
  recipes in c: the art of scientific computing,'' \emph{Camb. Univ. Press
  Camb}, 1992.

\bibitem{muhammad2021casia}
J.~Muhammad, Y.~Wang, C.~Wang, K.~Zhang, and Z.~Sun, ``Casia-face-africa: A
  large-scale african face image database,'' \emph{IEEE Transactions on
  Information Forensics and Security}, vol.~16, pp. 3634--3646, 2021.

\bibitem{USIT3}
C.~Rathgeb, A.~Uhl, P.~Wild, and H.~Hofbauer, ``Design decisions for an iris
  recognition sdk,'' in \emph{Handbook of Iris Recognition}, second
  edition~ed., ser. Advances in Computer Vision and Pattern Recognition,
  K.~Bowyer and M.~J. Burge, Eds.\hskip 1em plus 0.5em minus 0.4em\relax
  Springer, 2016.

\bibitem{zhao2015accurate}
Z.~Zhao and K.~Ajay, ``An accurate iris segmentation framework under relaxed
  imaging constraints using total variation model,'' in \emph{Proceedings of
  the IEEE international conference on computer vision}, 2015, pp. 3828--3836.

\bibitem{othman2016osiris}
N.~Othman, B.~Dorizzi, and S.~Garcia-Salicetti, ``Osiris: An open source iris
  recognition software,'' \emph{Pattern Recognition Letters}, vol.~82, pp.
  124--131, 2016.

\bibitem{wang2021nir}
C.~Wang, Y.~Wang, K.~Zhang, J.~Muhammad, T.~Lu, Q.~Zhang, Q.~Tian, Z.~He,
  Z.~Sun, Y.~Zhang, T.~Liu, W.~Yang, D.~Wu, Y.~Liu, R.~Zhou, H.~Wu, H.~Zhang,
  J.~Wang, J.~Wang, W.~Xiong, X.~Shi, S.~Zeng, P.~Li, H.~Sun, J.~Wang,
  J.~Zhang, Q.~Wang, H.~Wu, X.~Zhang, H.~Li, Y.~Chen, L.~Chen, M.~Zhang,
  Y.~Sun, Z.~Zhou, F.~Boutros, N.~Damer, A.~Kuijper, J.~Tapia, A.~Valenzuela,
  C.~Busch, G.~Gupta, K.~Raja, X.~Wu, X.~Li, J.~Yang, H.~Jing, X.~Wang,
  B.~Kong, Y.~Yin, Q.~Song, S.~Lyu, S.~Hu, L.~Premk, M.~Vitek, V.~Struc,
  P.~Peer, J.~N. Khiarak, F.~Jaryani, S.~S. Nasab, S.~N. Moafinejad, Y.~Amini,
  and M.~Noshad, ``Nir iris challenge evaluation in non-cooperative
  environments: Segmentation and localization,'' in \emph{2021 IEEE
  International Joint Conference on Biometrics (IJCB)}, 2021, pp. 1--10.

\bibitem{daugman1998recognizing}
J.~Daugman, ``Recognizing people by their iris patterns,'' \emph{Information
  Security Technical Report}, vol.~3, no.~1, pp. 33--39, 1998.


\bibitem{Ren2020DynamicGR}
M.~Ren, Y.~Wang, Z.~Sun, and T.~Tan, ``Dynamic graph representation for
  occlusion handling in biometrics,'' in \emph{AAAI Conference on Artificial
  Intelligence}, 2020.

\bibitem{Wu2015ALC}
X.~Wu, R.~He, Z.~Sun, and T.~Tan, ``A light cnn for deep face representation
  with noisy labels,'' \emph{IEEE Transactions on Information Forensics and
  Security}, vol.~13, pp. 2884--2896, 2015.

\bibitem{Zhang2018DeepFF}
Q.~Zhang, H.~Li, Z.~Sun, and T.~Tan, ``Deep feature fusion for iris and
  periocular biometrics on mobile devices,'' \emph{IEEE Transactions on
  Information Forensics and Security}, vol.~13, pp. 2897--2912, 2018.

\bibitem{Ren2019AlignmentFA}
M.~Ren, C.~Wang, Y.~Wang, Z.~Sun, and T.~Tan, ``Alignment free and distortion
  robust iris recognition,'' \emph{2019 International Conference on Biometrics
  (ICB)}, pp. 1--7, 2019.
  
\end{thebibliography}
\end{document}